\documentclass[letterpaper]{article} 
\usepackage{aaai23}  
\usepackage{times}  
\usepackage{helvet}  
\usepackage{courier}  
\usepackage[hyphens]{url}  
\usepackage{graphicx} 
\usepackage{natbib}  
\usepackage{caption} 
\usepackage{algorithm}
\usepackage{algorithmic}
\usepackage{amssymb}
\usepackage{amsthm} 
\usepackage{amsmath}
\usepackage{threeparttable}

\usepackage{amsfonts}
\usepackage{mathrsfs}

\usepackage{multirow}
\usepackage{makecell}
\usepackage{tablefootnote}
\usepackage{threeparttable}
\usepackage{color}

\newtheorem{theorem}{Theorem}
\newtheorem{corollary}{Corollary}
\newtheorem{lemma}{Lemma}
\newtheorem{definition}{Definition}
\newtheorem{remark}{Remark}
\usepackage{newfloat}
\usepackage{listings}
\DeclareCaptionStyle{ruled}{labelfont=normalfont,labelsep=colon,strut=off} 
\floatstyle{ruled}
\newfloat{listing}{tb}{lst}{}
\floatname{listing}{Listing}
\title{Stability-based Generalization Analysis for Mixtures of Pointwise and Pairwise Learning}
\author{
    Jiahuan Wang\textsuperscript{\rm 1},
    Jun Chen\textsuperscript{\rm 2},
    Hong Chen\textsuperscript{\rm 1, 3, 4,}\thanks{Corresponding author.},  
    Bin Gu\textsuperscript{\rm 5}, 
    Weifu Li\textsuperscript{\rm 1, 3, 4}, 
    Xin Tang\textsuperscript{\rm 6}\\
}
\affiliations{
    \textsuperscript{\rm 1}College of Science, Huazhong Agricultural University, Wuhan 430070, China\\
    \textsuperscript{\rm 2}College of Informatics, Huazhong Agricultural University, Wuhan 430070, China\\
    \textsuperscript{\rm 3}Engineering Research Center of Intelligent Technology for Agriculture, Ministry of Education, Wuhan 430070, China\\
    \textsuperscript{\rm 4}Key Laboratory of Smart Farming for Agricultural Animals, Wuhan 430070, China\\
    \textsuperscript{\rm 5}Mohamed bin Zayed University of Artificial Intelligence, Abu Dhabi, UAE\\
    \textsuperscript{\rm 6}Ping An Property \& Casualty Insurance Company, Shenzhen, China\\
   chenh@mail.hzau.edu.cn
}

\usepackage{bibentry}

\begin{document}

\maketitle

\begin{abstract}
Recently, some mixture algorithms of pointwise and pairwise learning (PPL) have been formulated by employing the hybrid error metric of “pointwise loss + pairwise loss” and have shown empirical effectiveness on feature selection, ranking and recommendation tasks.
However, to the best of our knowledge, the learning theory foundation of PPL has not been touched in the existing works. In this paper, we try to fill this theoretical gap by investigating the generalization properties of PPL. After extending the definitions of algorithmic stability to the PPL setting, we establish the high-probability generalization bounds for uniformly stable PPL algorithms. Moreover, explicit convergence rates of stochastic gradient descent (SGD) and regularized risk minimization (RRM) for PPL are stated by developing the stability analysis technique of pairwise learning. In addition, the refined generalization bounds of PPL are obtained by replacing uniform stability with on-average stability.
\end{abstract}

\section{Introduction}
There are mainly two paradigms to formulate machine learning systems including pointwise learning and pairwise learning. Usually, the former aims to train models under the error metric associated with single sample, while the latter concerns the relative relationships between objects measured by the loss related to the pair of samples. Besides wide applications, the theoretical foundations of the above paradigms have been well established from the viewpoint of statistical learning theory, e.g., pointwise stability analysis \cite{bousquet2002stability, london2016stability, sun2021stability}, pairwise stability analysis \cite{agarwal2009generalization, lei2020sharper, lei2021generalization}, and uniform convergence analysis  \cite{clemenccon2008ranking, rejchel2012ranking, cao2016generalization, ying2016stochastic, ying2016online}.

It is well known that pointwise (or pairwise) learning enjoys certain advantages and limitations for real-world data analysis. 
For the same number of samples, pointwise learning has computation feasibility due to its low model complexity, while pairwise learning can mine valuable information in terms of the intrinsic relationship among samples.  
As illustrated by \citet{wang2016ppp}, the degraded performance may occur for the pointwise learning with the ambiguity of some labels and for the pairwise learning as samples in different categories have similar features. 
Therefore, it is natural to consider the middle modality of the above paradigms to alleviate  their drawbacks. 
Along this line, some learning algorithms have been proposed under the pointwise and pairwise learning (PPL) framework, where the pointwise loss and the pairwise loss are employed jointly \cite{liu2015pairwise, wang2016ppp, lei2017alternating, zhuo2022learning, wang2022mp2}. 
In PPL, its pointwise part concerns the fitting ability to empirical observations and its pairwise part addresses the stability or robustness of learning models \cite{liu2015pairwise}. 
While the studies on algorithmic design and applications are increasing, there are far fewer results to investigate the generalization ability of PPL in theory.
\begin{table*}[h]
    \begin{center}
        \renewcommand\arraystretch{1.5}
        \begin{threeparttable}
            \begin{tabular}{ccccc}
            \hline
            Objective function & Reference & Optimization & Application & Generalization analysis\\
            \hline
            \multirow{4}{*}{\thead{Pointwise loss +Pairwise loss\\ +Regularization}} & \citet{liu2015pairwise} & ACG & Feature Selection & No\\
            & \citet{wang2016ppp} & ADMM & Image Classification &No\\
            & \citet{lei2017alternating} & SGD & Ranking &No\\
            & \citet{wang2022mp2} & SGD & Recommendation &No\\
            \hline
            \end{tabular}
        \end{threeparttable}
        \caption{Summary of pointwise and pairwise learning (ACG: Accelerated Proximal Gradient; ADMM: Alternating Direction Method of Multipliers).}
        \label{Comparison1}
        \end{center}
\end{table*}

As one of the main routines of learning theory analysis, the algorithmic stability tools have advantages in some aspects, such as dimensional independence and adaptivity for broad learning paradigms \cite{bousquet2002stability,shalev2010learnability,hardt2016train,feldman2018generalization,feldman2019high}.
Specially, the stability and generalization have been well understood recently for stochastic gradient descent (SGD) and regularized risk minimization (RRM) under both the pointwise learning \cite{hardt2016train, lei2020fine} and the pairwise learning setting \cite{lei2020sharper, lei2021generalization}.  
Inspired by the recent progress, in this paper, we try to fill this theoretical gap of PPL by establishing its generalization bounds in terms of the algorithmic stability technique. To the best of our knowledge, this is the first theoretical understanding of generalization properties for PPL.

The main work of this paper is two-fold: 
One is to establish a relationship between uniform stability and estimation error for the mixture setting, which can be considered as natural extension of the related results in \cite{lei2020sharper, lei2021generalization}. 
The other is to characterize the stability-based generalization  bounds for some PPL algorithms (i.e. SGD and RRM) under mild conditions.

\section{Related Work}
To better evaluate our theoretical results, we review the related works on PPL and generalization analysis.

{\bf{Pointwise and pairwise learning (PPL).}}
In recent years, some algorithms of PPL have been designed for learning tasks such as feature selection, image classification, ranking, and recommendation systems.
\citet{liu2015pairwise} proposed a pairwise constraint-guided sparse learning method for feature selection, where the pairwise constraint is used for improving robustness. 
For image classification tasks, \citet{wang2016ppp} designed a novel joint framework, called pointwise and pairwise image label prediction, to predict both pointwise and pairwise labels and achieved superior performance. For emphasis selection tasks, \citet{huang2020ernie} employed a pointwise regression loss and a pairwise ranking loss simultaneously to fit models.
Recently, \citet{lei2017alternating} proposed an alternating pointwise-pairwise ranking to improve decision performance. 
\citet{zhuo2022learning} and \citet{wang2022mp2} formulated hybrid learning models for the recommendation systems. Although the empirical effectiveness has been validated for the above PPL algorithms, their theoretical foundations (e.g., generalization guarantee) have not been investigated before. 
To further highlight the gap in generalization analysis, we summarize the basic properties of PPL models in Table \ref{Comparison1}.

{\bf{Generalization analysis.}}
From the viewpoint of statistical learning theory, generalization analysis is crucial since it provides the statistical theory support for the empirical performance of trained models. 
Usually, model training is a process of calculating loss based on data and then seeking an optimal function in the predetermined hypothetical function space through an optimization algorithm. 
Naturally, the generalization performance of learning systems can be investigated  from the perspectives of hypothetical function space \cite{smale2007learning,yin2019rademacher,lei2021learning,WangCZXGC20} and data \cite{bousquet2002stability,elisseeff2005stability,shalev2010learnability}, respectively. The former is often called uniform convergence analysis and the latter is realized by stability analysis. In essential, the uniform convergence analysis considers the capacity of hypothesis space (e.g., via VC-dimension \cite{vapnik1994measuring}, covering numbers \cite{zhou2002covering, ChenWDH17,ChenWZDH21}, Rademacher complexity \cite{yin2019rademacher}), while the stability analysis concerns the change of model parameters caused by the change of training data \cite{bousquet2002stability,lei2021generalization}. 
Algorithmic stability has shown remarkable effectiveness in deriving dimension-independent generalization bounds for wide learning frameworks.
A classic framework for stability analysis is developed by \citet{bousquet2002stability}, in which the uniform stability and hypothesis stability are introduced. 
Subsequently, the uniform stability measure was extended to study stochastic algorithms \cite{elisseeff2005stability, hardt2016train} and inspired several other stability concepts including uniform argument stability \cite{liu2017algorithmic}, locally elastic stability \cite{deng2021toward},  on-average loss stability \cite{lei2020sharper,lei2020fine,lei2021generalization} and on-average argument stability \cite{shalev2010learnability, lei2021generalization}.

From the lens of learning paradigms, generalization guarantees have been established for various pointwise learning algorithms
\cite{bousquet2002stability, london2016stability,hardt2016train,lei2020fine,sun2021stability,klochkov2021stability} and  pairwise learning models \cite{agarwal2009generalization, lei2020sharper, lei2021generalization,DBLP:conf/nips/YangLWYY21}.
Therefore, it is natural to explore the generalization properties of PPL by the means of the stability analysis technique. 

\section{Preliminaries}
This section introduces the problem formulation of PPL and the definitions of algorithmic stability. 

\subsection{Pointwise and Pairwise Learning}

Consider a training dataset $S:=\{z_i\}_{i=1}^n$, where each $z_{i}$ is independently drawn from a probability measure $\rho$ defined over a sample space $\mathcal{Z}=\mathcal{X} \times \mathcal{Y}$. Here, $\mathcal{X} \subset \mathbb{R}^{d}$ is an input space of dimension $d$ and $\mathcal{Y} \subset \mathbb{R}$ is an output space.
Let $\mathcal{W}$ be a given parameter space of learning models. The goal of pointwise learning is to find a parameter $\mathbf{w}$ based model such that  the population risk (or expected risk), defined as
$$
R^{point}(\mathbf{w})=\mathbb{E}_{z}[f(\mathbf{w}; z)],
$$ 
is as small as possible, 
where $f: \mathcal{W} \times \mathcal{Z} \rightarrow [0, \infty)$ is a pointwise loss  and $\mathbb{E}_{z}$ denotes the expectation with respect to ${z}\sim\rho$. For brevity, we also use $\mathbf{w}$ to denote the parameter $\mathbf{w}$ based model in the sequel. 

However, we can't get the minimizer of $R^{point}(\mathbf{w})$ directly since the intrinsic distribution $\rho$ is unknown. As a natural surrogate, for algorithmic design,  we often  consider the corresponding empirical risk defined as
$$
R_{S}^{point}(\mathbf{w})=\frac{1}{n} \sum_{i=1}^{n} f\left(\mathbf{w} ; z_{i}\right).
$$

Unlike the pointwise learning, the pairwise learning model $\mathbf{w}$ is  measured by  
$$
R^{pair}(\mathbf{w})=\mathbb{E}_{z, \tilde{z}}[g(\mathbf{w} ; z, \tilde{z})],
$$ 
where $g: \mathcal{W} \times \mathcal{Z} \times \mathcal{Z} \rightarrow [0, \infty)$ is a pairwise loss function and $\mathbb{E}_{z, \tilde{z}}$ denotes the expectation with respect to ${z, \tilde{z}}\sim\rho$. 
In pairwise learning models, $R^{pair}(\mathbf{w})$ is approximately characterized by the empirical risk
$$
R_{S}^{pair}(\mathbf{w})=\frac{1}{n(n-1)} \sum_{i, j \in[n]: i \neq j} g\left(\mathbf{w} ; z_{i}, z_{j}\right),
$$
where $z_i,z_j\sim\rho$ and $[n]:=\{1, \ldots, n\}$. 

In this paper, we consider a mixture paradigm of pointwise learning and pairwise learning, called pointwise and pairwise learning (PPL). The population risk of $\mathbf{w}$ in PPL is 
\begin{equation*}
    R(\mathbf{w})=\tau R^{point}(\mathbf{w})+(1-\tau)R^{pair}(\mathbf{w}),
\end{equation*}

where $\tau\in[0,1]$ is a tuning parameter. Given training set $S$, the corresponding empirical version of $R(\mathbf{w})$ is 
\begin{equation}
	R_{S}(\mathbf{w})=\tau R_{S}^{point}(\mathbf{w})+(1-\tau)R_{S}^{pair}(\mathbf{w}).
	\label{pointwise and pairwise empirical risk}
\end{equation}
 
For brevity, $A(S)$ denotes the derived model by applying algorithm $A$ (e.g., SGD and RRM) on $S$. In the process of training and adjustment of parameters, the output model $A(S)$ can be a small empirical risk since we often can fit training examples perfectly. However, the empirical effectiveness of $A(S)$ can not assure the small population risk. In statistical learning theory, the difference between the population risk and empirical risk  
\begin{equation}
    R(\mathbf{w})-R_{S}(\mathbf{w})
\end{equation}
is called the generalization error of learning model $\mathbf{w}$.
It is key concern of this paper to bound this gap in theory. 

\subsection{Algorithmic stability}
Algorithmic stability is an important concept in statistical learning,  which measures the sensitivity of an algorithm to the perturbation of training sets. This paper focuses on the analysis techniques associated with the algorithmic uniform stability \cite{bousquet2002stability, elisseeff2005stability, agarwal2009generalization, hardt2016train}, on-average loss stability \cite{lei2020sharper,lei2020fine,lei2021generalization}, and on-average argument stability \cite{shalev2010learnability,lei2021generalization}.
To match  the generalization analysis of PPL algorithms, we firstly extend the definitions of algorithmic stability (e.g., the pointwise uniform stability  \cite{bousquet2002stability} and pairwise uniform stability \cite{lei2020sharper}) to the PPL setting.

Let $S=\left\{z_{1}, \ldots, z_{n}\right\}$ and  $S^{\prime}=\left\{z_{1}^{\prime}, \ldots, z_{n}^{\prime}\right\}$ be independently drawn from $\rho$. For any $i<j, i,j\in[n]$, denote 
\begin{equation}\label{Si}
	S_{i}=\left\{z_{1}, \ldots,z_{i-1},z_{i}^{\prime},z_{i+1},\ldots, z_{n}\right\}
\end{equation}
and
\begin{align}\label{Sij}
	S_{i,j}=\left\{z_{1}, \ldots,z_{i-1},z_{i}^{\prime},z_{i+1},\ldots,z_{j-1},z_{j}^{\prime},z_{j+1},\ldots, z_{n}\right\}.
\end{align}

\begin{definition}\label{pointwise and pairwise Uniform Stability}	(PPL Uniform Stability).
	Assume that $f(\cdot; z)$ is a pointwise loss function and $g(\cdot; z, \tilde{z})$ is a pairwise loss function. We say $A: \mathcal{Z}^{n} \mapsto \mathcal{W}$ is PPL $\gamma$-uniformly stable, if for any training datasets $S, S_{i} \in \mathcal{Z}^{n}$
	\begin{equation*}
	    \max\left\{U_{point}, U_{pair}\right\}\leq \gamma, \forall i\in[n],
	\end{equation*}
	where $U_{point}=\sup _{z\in \mathcal{Z}}\left|f(A(S) ; z )-f\left(A\left(S_{i}\right) ; z\right)\right|$ and $U_{pair}=\sup _{z, \tilde{z} \in \mathcal{Z}}\left|g(A(S) ; z, \tilde{z})-g\left(A\left(S_{i}\right) ; z, \tilde{z}\right)\right|$.
\end{definition}

\begin{remark} 
Denote  $$\ell(A(S_i);z,\tilde{z}) :=\tau f(A(S_i);z)+(1-\tau)g(A(S_i);z,\tilde{z})$$
for simplicity, and call it as the PPL loss function. Then, we can define the weaker stability measure by replacing the maximum in Definition \ref{pointwise and pairwise Uniform Stability} with
\begin{equation*}
    \sup _{z, \tilde{z} \in \mathcal{Z}}\left|\ell(A(S); z, \tilde{z})-\ell\left(A\left(S_{i}\right); z, \tilde{z}\right)\right|\leq \gamma.
\end{equation*}
Following the mixture stability associated with $\tau$, we can also get the similar generalization results as Theorems 1 and 2 in the next section.
\end{remark}

We then introduce the definitions of PPL on-average loss stability and PPL on-average argument stability described as follows.
\begin{definition}
	\label{Pointwise and pairwise On-average Stability}
	(PPL On-average Loss Stability).
Let $f(\cdot ; z)$ be a pointwise loss function and let $g(\cdot ; z, \tilde{z})$ be a pairwise loss function. We say $A: \mathcal{Z}^{n} \mapsto \mathcal{W}$ is  PPL $\gamma$-on-average loss stable if, for any training datasets $S, S_{i}, S_{i,j} \in \mathcal{Z}^{n}$,
	\begin{equation*}
	    \max\left\{V_{point}, V_{pair}\right\}\leq \gamma, \forall i<j\in[n],
	\end{equation*}
	where 
	$$
	V_{point}=\frac{1}{n} \sum_{i\in[n]} \mathbb{E}_{S, S^{\prime}}\left[f\left(A\left(S_{i}\right) ; z_{i}\right)-f\left(A(S) ; z_{i}\right)\right]
	$$
	and 
	$$
	V_{pair}=\frac{\sum_{i, j \in[n]} \mathbb{E}_{S, S^{\prime}}\left[g\left(A\left(S_{i, j}\right) ; z_{i}, z_{j}\right)-g\left(A(S) ; z_{i}, z_{j}\right)\right]}{n(n-1)}.
	$$
	\end{definition}
	
\begin{remark}
Definition \ref{Pointwise and pairwise On-average Stability} is built  by combining the on-average loss stability for pointwise learning \cite{shalev2010learnability,lei2020fine} with the one for pairwise learning \cite{lei2020sharper,lei2021generalization}. The requirement of  PPL on-average loss stability is milder than the PPL uniform stability, where the stabilization is not measured by the changes of all samples but the mean of training sets.
\end{remark}

\begin{definition}\label{ppl argument}
    (PPL On-average Argument Stability).
    We say $A: \mathcal{Z}^{n} \mapsto \mathcal{W}$ is PPL $\ell_1$ $\gamma$-on-average argument stable if, for any training datasets $S, S_{i} \in \mathcal{Z}^{n}$,
    \begin{equation}\label{l1}
        \max\left\{\mathbb{E}_{S, S^{\prime},A}[H_{point}],\mathbb{E}_{S, S^{\prime},A}[H_{pair}]\right\}\leq\gamma,
    \end{equation}
    where $H_{point}=\frac{1}{n}\sum_{i=1}^{n}\|A(S)-A(S_i)\|_2$ and $H_{pair}=\frac{1}{n(n-1)}\sum_{i,j\in[n]:i\neq j}\|A(S)-A(S_{i,j})\|_2$.\\
    We say $A$ is PPL $\ell_2$ $\gamma$-on-average argument stable if, for any training datasets $S, S_{i} \in \mathcal{Z}^{n}$,
    \begin{equation*}
        \mathbb{E}_{S, S^{\prime},A}\Big[\frac{1}{n}\sum_{i=1}^{n}\|A(S)-A(S_i)\|_2^2\Big]\leq\gamma^2.
    \end{equation*}
\end{definition}

\begin{remark}
    The $\ell_2$ on-average argument stability of PPL is similar with that of pointwise learning \cite{lei2020fine} and pairwise learning \cite{lei2021generalization}. Differently from Definition \ref{pointwise and pairwise Uniform Stability}, which relied on the drift of loss functions, Definition \ref{Pointwise and pairwise On-average Stability}, \ref{ppl argument} measures the stability in terms of the changes of the model $A(S)$.
\end{remark}

\section{Main Results}
In this section, we present our main results on the generalization bounds of PPL algorithms based on uniform stability and on-average stability. 

\subsection{Uniform stability-based generalization}
This subsection establishes the relationship between the generalization ability and the uniform stability for PPL. 
In the sequel, $e$ represents the base of the natural logarithm, $\lceil a \rceil$ means the smallest integer which is no less than $a$.
\emph{Supplementary Material B.1} provides the detailed proof of the following theorem.

\begin{theorem}\label{Generalization by algorithmic stability}
Let $A: \mathcal{Z}^{n} \mapsto \mathcal{W}$ be PPL $\gamma$-uniformly stable. \\
Assume that
$$\max_{z, \tilde{z} \in \mathcal{Z}}\{\left|\mathbb{E}_{S}[f(A(S); z)]\right|, \left|\mathbb{E}_{S}[g(A(S); z, \tilde{z})] \right|\}\leq M$$ for some positive constant $M$.
Then, for all $\tau \in[0,1]$ and $\delta \in(0,1 / e)$, we have, with probability $1-\delta$,  
	\begin{align*}
		&\left|R_{S}(A(S))-R(A(S))\right|\\
		\leq & (4-2\tau)\gamma+e\left(4M(4-3\tau)n^{-\frac{1}{2}} \sqrt{\log (e / \delta)}\right.\\
		& + \left.24 \sqrt{2}(2-\tau) \gamma\left\lceil\log _{2}(n)\right\rceil \log (e / \delta)\right).
	\end{align*}
\end{theorem}

\begin{remark}
	 Theorem \ref{Generalization by algorithmic stability} is a high-probability generalization bound for uniformly stable PPL algorithms, motivated by the recent analyses in the pointwise learning \cite{hardt2016train} and the pairwise learning \cite{lei2020sharper, lei2021generalization}. Similar to \citet{lei2020sharper}, the error estimations of the pointwise part and the pairwise part of PPL are obtained by applying and developing the concentration inequality \cite{bousquet2020sharper}. Ignoring the constants (e.g. $M$, $\tau$, $\log (e / \delta)$), we can get the convergence order $O(n^{-\frac{1}{2}} + \gamma \log_2n)$ from Theorem \ref{Generalization by algorithmic stability}. Due the generality and flexibility of PPL induced by $\tau\in[0,1]$, the derived connection, between uniform stability and generalization, contains the previous results for pairwise learning \cite{lei2020sharper} as special example.
\end{remark}

\begin{remark}
    To better understand the stability-based generalization bound, we summarize the main results about the relationships between generalization and various definitions of algorithmic stability in Table \ref{Comparison2}. 
    In this table, for feasibility, we denote the generalization error as
    $$Gen := R(A(S))-R_{S}(A(S))$$ 
    and denote the expected generalization error as
    $$\mathbb{E}_{Gen} := \mathbb{E}_{S,A}[R(A(S))-R_{S}(A(S))].$$
    Table \ref{Comparison2} demonstrates our characterized relations are comparable with the existing results. 
\end{remark}

\begin{table*}[hh]
    \begin{center}
        \renewcommand\arraystretch{1.5}
            \begin{tabular}{c|ccc}
            \hline
            Type & Reference & Algorithmic stability ($\gamma$) & Relationship \\
            \hline
            \multirow{7}{*}{\thead{Pointwise\\ Learning}} & \citet{bousquet2002stability} & Uniform Stability & $\mathbb{E}_{Gen}\leq \gamma$.\\
            & \citet{bousquet2002stability} & Hypothesis Stability & $Gen=O(\sqrt{\gamma}).$\\
            & \citet{shalev2010learnability} & On-average Stability & $\mathbb{E}_{Gen}\leq \gamma$.\\
            & \citet{hardt2016train} & Uniform Stability & $\mathbb{E}_{Gen}\leq \gamma$. \\
            & \citet{feldman2018generalization} & Uniform Stability & $Gen=O(\sqrt{\gamma }+n^{-\frac{1}{2}})$.\\
            & \citet{feldman2019high} & Uniform Stability & $Gen=O\left(\gamma(\log_{2}n)^2+n^{-\frac{1}{2}}\right)$.\\
            & \citet{lei2020fine} & On-average Stability & $\mathbb{E}_{Gen}\leq L\gamma$. \\
            \hline
            \multirow{3}{*}{\thead{Pairwise \\Learning}} & \citet{lei2020sharper} & Uniform Stability & $Gen= 4\gamma+O(n^{-\frac{1}{2}}+\gamma \log_{2} n)$.\\
            & \citet{lei2021generalization} & $\ell_1$ On-average Argument Stability & $\mathbb{E}_{Gen}= O(\gamma^2+\gamma/n)$.\\
            & \citet{lei2021generalization} & $\ell_2$ On-average Argument Stability & $\mathbb{E}_{Gen}= O(\gamma^2+\gamma/\sqrt{n})$.\\
            \hline
            \multirow{4}{*}{\thead{PPL}} & \multirow{2}{*}{Ours (Theorem \ref{Generalization by algorithmic stability})} & \multirow{2}{*}{\thead{Uniform Stability}} & $Gen= (4-2\tau)\gamma+O((4-3\tau)n^{-\frac{1}{2}}$\\& & &$+(2-\tau)\gamma \log_{2} n)$.\\
            & Ours (Theorem \ref{connection}) & On-average Loss Stability & $\mathbb{E}_{Gen}\leq \gamma$.\\
            & Ours (Corollary \ref{Thl1})& $\ell_1$ On-average Argument Stability & $\mathbb{E}_{Gen}\leq L\gamma$.\\
            & Ours (Theorem \ref{Thl2})& $\ell_2$ On-average Argument Stability & $\mathbb{E}_{Gen}=O(\frac{1}{\gamma}+(2-\frac{3}{2}\tau)\gamma)$.\\
            \hline
            \end{tabular}
        \caption{Summary of stability-based generalization bounds ($\gamma$-stability parameter; $Gen$-generalization error; $\mathbb{E}_{Gen}$-expected generalization error).}
        \label{Comparison2}
    \end{center}
\end{table*}

\subsection{Generalization bounds of SGD for PPL}
As a popular computing strategy, SGD has been employed for PPL as shown in Table \ref{Comparison1}. The SGD for PPL can be regarded as an elastic net version of pointwise SGD and pairwise SGD, which involves the gradients of the pointwise loss function $f$ and the  pairwise loss function $g$. At the $t$-th iteration, $\left(i_{t}, j_{t}\right)$ is taken from the uniform distribution over $[n]$ randomly, which requires $i_t\neq j_t$. The SGD for the PPL model is updated by
\begin{equation}\label{SGD}
	\mathbf{w}_{t+1}=\mathbf{w}_{t}-\eta_{t}\left(\tau \nabla f\left(\mathbf{w}_{t} ; z_{i_{t}}\right)+(1-\tau)\nabla g\left(\mathbf{w}_{t} ; z_{i_{t}}, z_{j_{t}}\right)\right),   
\end{equation}
where $\left\{\eta_{t}\right\}_{t}$ is a step size sequence, and $\nabla f\left(\mathbf{w}_{t} ; z_{i_{t}}\right)$ and $\nabla g\left(\mathbf{w}_{t} ; z_{i_{t}}, z_{j_{t}}\right)$ denote the subgradients of $f\left(\cdot; z_{i_{t}}\right)$ and $g\left(\cdot ; z_{i_{t}}, z_{j_{t}}\right)$ at $\mathbf{w}_{t}$, respectively.

To bound the gradient update process of SGD, it is necessary to presume some properties of the loss functions. For brevity, we just recall some conditions for pointwise loss function \cite{hardt2016train,lei2020fine} since the definitions of pairwise setting are analogous.

\begin{definition}
    A loss function $f: \mathcal{W} \times \mathcal{Z} \rightarrow [0, \infty)$ is $\sigma$-strongly convex if
    $$
    f(u) \geq f(v)+\langle\nabla f(v), u-v\rangle+\frac{\sigma}{2}\|u-v\|_2^{2}, \forall u, v \in \mathcal{W}.
    $$
  Specially, $f$ is convex if $\sigma = 0$.
\end{definition}
Clearly, a strongly convex loss function must be convex, but the contrary may not be true. It is well known that convexity is crucial for some optimization analyses of learning algorithms \cite{hardt2016train,harvey2019tight}. 

\begin{definition}
    A loss function $f: \mathcal{W} \times \mathcal{Z} \rightarrow [0, \infty)$ is L-Lipschitz if
    $$
    |f(u)-f(v)| \leq L\|u-v\|_2, \forall u, v \in \mathcal{W}.
    $$
\end{definition}
The above inequality is equivalent to the gradient boundedness of $f$, i.e. $\|\nabla f(x)\|_2 \leq L$.
Thus, the $L$-Lipschitz continuity assures the boundedness of the gradient update.

\begin{definition}
    A loss function $f: \mathcal{W} \times \mathcal{Z} \rightarrow [0, \infty)$ is $\beta$-smooth if
    $$
    \|\nabla f(u)-\nabla f(v)\|_2 \leq \beta\|u-v\|_2, \forall u, v \in \mathcal{W}.
    $$
  \end{definition}
Following the steps in \cite{hardt2016train,lei2020sharper}, we can verify that the gradient update is non-expansive when $f$ is convex and $\beta$-smooth. 

Now we present the generalization bounds of SGD for PPL. The proof is given in \emph{Supplementary Material B.2}.
\begin{theorem}\label{Th2}
	Suppose for any $z,\tilde{z}\in \mathcal{Z}$, $f\left(\mathbf{w} ; z\right)$ and $g\left(\mathbf{w} ; z, \tilde{z}\right)$ are convex, $\beta$-smooth and $L$-Lipschitz with respect to $\mathbf{w}\in\mathcal{W}$.
	Without loss of generality, let $S$ and $S^{\prime}$ be different only in  the last example. If $\eta_{t} \leq 2 / \beta$, then SGD  for PPL  with $t$ iterations is PPL $\gamma$-uniformly stable with
	\begin{equation}\label{stab}
	    \gamma \leq 2 L^2 \sum_{k=1}^{t} \eta_{k} \mathbb{I}\left[i_{k}=n\right]+2 L^2 (1-\tau)\sum_{k=1}^{t} \eta_{k} \mathbb{I}\left[j_{k}=n\right],
	\end{equation}
	where $\mathbb{I}[\cdot]$ is the indicator function.
	Let $\{\mathbf{w}_t\}$, $\{\mathbf{w}_t^{\prime}\}$ be generated by SGD on $S$ and $S^{\prime}$ with $\eta_t=\eta$. Then, for all $\delta \in(0,1 / e)$, the following inequality holds with probability $1-\delta$
	\begin{align*}
	&\left\|\mathbf{w}_{t+1}-\mathbf{w}_{t+1}^{\prime}\right\|_{2}\\ 
	\leq&2L\eta(2-\tau)\Big(\frac{t}{n}+\log(1/\delta)+\sqrt{2 n^{-1} t  \log (1 / \delta)}\Big).
	\end{align*}
\end{theorem}

Theorem \ref{Th2} characterizes the impact of the change of training set on the training loss and the model parameter, which extends the previous related results of pointwise (or pairwise) SGD to the general PPL setting.

We now apply Theorem \ref{Generalization by algorithmic stability} with $A(S)=\mathbf{w}_T$ where $T$ is the index of the last iteration and the uniform stability bounds in \eqref{stab} to derive the following result.
\begin{corollary}\label{th3}
   Suppose that $f\left(\mathbf{w} ; z\right)$ and $g\left(\mathbf{w} ; z, \tilde{z}\right)$ are convex, $\beta$-smooth and $L$-Lipschitz with respect to $\mathbf{w}$, and
   $$\max_{z, \tilde{z} \in \mathcal{Z}}\{\left|\mathbb{E}_{S}[f(\mathbf{w}_T; z)]\right|,\left|\mathbb{E}_{S}[g(\mathbf{w}_T ; z, \tilde{z})]\right|\}\leq M$$ for some positive constant $M$, where $\mathbf{w}_T$ is produced by SGD \eqref{SGD} at $T$-th iteration with $\eta_{t}\equiv c/\sqrt{T} \leq 2 / \beta$. Then, for any $\delta \in(0,1 / e)$, the following inequality holds with probability $1-\delta$
   	\begin{eqnarray*}
		\left|R_{S}(\mathbf{w}_T)-R(\mathbf{w}_T)\right| 
		=O\Big(\sqrt{\frac{\log (\frac1\delta)}{n}}+\frac{\sqrt{T}}{n}\log_{2}n\log (\frac1\delta)\Big)\\
		+O\Big(T^{-\frac12}\log_{2}n\log^2 (\frac1\delta)+n^{-\frac12}\log_{2}n\log^{\frac32} (\frac1\delta)\Big).
	\end{eqnarray*}
\end{corollary}

\begin{remark}
    Let 
    \begin{equation}\label{wr*}
       \mathbf{w}_{R}^{*}=\arg \min _{\mathbf{w} \in \mathcal{W}} R(\mathbf{w}).
    \end{equation}
    The excess risk of SGD for PPL is defined as 
    \begin{align*}
        R(\mathbf{w}_T)-R(\mathbf{w}_{R}^{*})=&\left[R(\mathbf{w}_T)-R_S(\mathbf{w}_T)\right]+\left[R_S(\mathbf{w}_T)\right.\\
        &\left.-R_S(\mathbf{w}_{R}^{*})\right]+\left[R_S(\mathbf{w}_{R}^{*})-R(\mathbf{w}_{R}^{*})\right],
    \end{align*}
    where the first term and the second term of right side are called estimation error (or generalization error) and  optimization error, respectively. Theorem 3 provides the upper bound of the first term and  the results of \cite{harvey2019tight} imply the bound  $O(T^{-\frac{1}{2}}\log_{2}T)$ for the optimization error. For the third term $R_S(\mathbf{w}_{R}^{*})-R(\mathbf{w}_{R}^{*})$, we can bound it by Bernstein's inequality. When $T=O(n)$, with probability $1-\delta$ we have
    \begin{equation*}
        R(\mathbf{w}_T)-R(\mathbf{w}_{R}^{*})=O(n^{-\frac{1}{2}}\log_{2}n),
    \end{equation*}
  which is comparable with the convergence analysis of SGD for pairwise learning \cite{lei2020sharper}. 
\end{remark}

\subsection{Generalization bounds of RRM for PPL}
Let $r: \mathcal{W}\rightarrow [0, \infty)$ be a regularization term for achieving sparsity or preventing over-fitting of learning algorithms associated with  $R_S(w)$ defined in \eqref{pointwise and pairwise empirical risk}. The RRM for PPL aims to search the mininizer of 
\begin{equation}
	\label{RRM}
		F_S(\mathbf{w}):=R_S(\mathbf{w}) + r(\mathbf{w})
\end{equation}
over $\mathbf{w}\in\mathcal{W}$. 
Let
\begin{equation}\label{w*}
    \mathbf{w}^{*}=\arg \min _{\mathbf{w} \in \mathcal{W}}[R(\mathbf{w})+r(\mathbf{w})]
\end{equation}
and
\begin{equation}\label{as}
    A(S)=\arg \min _{\mathbf{w} \in \mathcal{W}} F_{S}(\mathbf{w}).
\end{equation}
We can verify the uniform stability of PPL with a strongly convex loss function, which is proved in \emph{Supplementary Material B.3}. 

\begin{lemma}\label{stable}
	Assume that $A$ is defined by \eqref{as}. Suppose $F_{S}$ is $\sigma$-strongly convex and the pointwise loss function $f\left(\cdot; z\right)$ and pairwise loss function $g\left(\cdot; z, \tilde{z}\right)$ are both L-Lipschitz. Then, $A$ is $\frac{4 L^{2}}{n \sigma}(2-\tau)$-uniformly stable.
\end{lemma}
The following lemma shows the distance between the empirical optimal solution (the best algorithm learned in the training set) and the theoretically optimal solution in expectation.
\begin{lemma}\label{juli}
 Assume that $F_{S}$ is $\sigma$-strongly convex. If the algorithm $A$ defined in \eqref{as} is PPL $\gamma$-uniformly stable, then $$\mathbb{E}_S\left\|A(S)-\mathbf{w}^{*}\right\|_2^{2} \leq  4\gamma(2-\tau) / \sigma.$$
\end{lemma}

A mixed version of Bernstein's inequality from \cite{hoeffding1994probability,pitcan2017note,lei2020sharper} is also introduced here, which is used in our error analysis.  
\begin{lemma}\label{bern}
    Assume that
    $$
   \min_{z, \tilde{z} \in \mathcal{Z}}\{f(\mathbf{w}^*; z), g(\mathbf{w}^*; z, \tilde{z})\}\geq 0,
    $$
    $$
    \max_{z, \tilde{z} \in \mathcal{Z}}\{f(\mathbf{w}^* ; z), g(\mathbf{w}^* ; z, \tilde{z})\}\leq b
    $$
    for some constants $b, \theta>0$, and 
    \begin{equation*}
        \max\{Var[f\left(\mathbf{w}^{*} ; Z\right)], Var[g(\mathbf{w}^{*} ; Z, \tilde{Z})]\}\leq \theta,
    \end{equation*}
    where $Var(a)$ denotes the variance of $a$ and $\mathbf{w}^*$ is defined by \eqref{w*}.
    Then, for any $\delta\in (0, 1)$, 
	with probability at least $1 - \delta$ we have
	\begin{eqnarray*}
        \big|R(\mathbf{w}^*) - R_S(\mathbf{w}^*)\big|
        \leq \frac{2(1-\tau)b\log(1/\delta)}{3\lfloor n/2 \rfloor}+ \frac{2\tau b\log(1/\delta)}{3\lfloor n \rfloor} \\
        + (1-\tau) \sqrt{\frac{2\theta\log(1/\delta)}{\lfloor n/2 \rfloor}}+ \tau \sqrt{\frac{2\theta\log(1/\delta)}{\lfloor n \rfloor}},
	\end{eqnarray*}
	where $\lfloor a \rfloor$ is the biggest integer no more than $a$.
\end{lemma}

Next, we firstly derive the upper bounds of the pointwise loss function and the pairwise loss function, and then apply Theorem \ref{Generalization by algorithmic stability} to get the generalization bounds for PPL with strongly convex objective functions. 
\begin{theorem}\label{rrm}
Assume that $F_S(\mathbf{w})$ is $\sigma$-strongly convex,  $f\left(\cdot; z\right)$ and  $g\left(\cdot; z, \tilde{z}\right)$ are both $L$-Lipschitz. Under the assumptions of  Lemma \ref{bern}, for the RRM algorithm $A$ defined by \eqref{as} and any $\delta\in (0, 1/e)$, with probability $1-\delta$ we have
\begin{align*}
	&\left|R_{S}(A(S))-R(A(S))\right| \\
   \leq&\frac{2b\log(1/\delta)}{3\lfloor n \rfloor} +\sqrt{\frac{2\theta\log(1/\delta)}{\lfloor n \rfloor}}+\frac{8 L^{2}}{n \sigma}(2-\tau)^2\\
   &+e\left(\frac{16 L^{2}}{n \sigma}(2-\tau)(4-3\tau) \sqrt{\log (e / \delta)}\right.\\
   &+ \left.\frac{96\sqrt{2} L^{2}}{n \sigma}(2-\tau)^2\left\lceil\log _{2}(n)\right\rceil \log (e / \delta)\right).
\end{align*}
\end{theorem}

\begin{remark}
    Note that the excess risk 
    \begin{align*}
        &R(A(S))-R(\mathbf{w}^{*}_R)\\
        =&\left[R(A(S))-R_S(A(S))\right]+\left[R_S(A(S)-R_S(\mathbf{w}^{*}_R)\right]\\
        &+\left[R_S(\mathbf{w}^{*}_R)-R(\mathbf{w}^{*}_R)\right]
        \\
        =&\left[R(A(S))-R_S(A(S))\right]+\left[R_S(\mathbf{w}^{*}_R)-R(\mathbf{w}^{*}_R)\right]\\
        &+\left[F_S(A(S)-F_S(\mathbf{w}^{*}_R)\right]+r(\mathbf{w}^{*}_R)-r(A(S))\\
        \leq &\left[R(A(S))-R_S(A(S))\right]+\left[R_S(\mathbf{w}^{*}_R)-R(\mathbf{w}^{*}_R)\right]\\
        &+r(\mathbf{w}^{*}_R)-r(A(S)),
    \end{align*}
    where $\mathbf{w}^{*}_{R}$ is defined by \eqref{wr*}.
    Following the similar proof strategy of Theorem 4,  we derive    
    \begin{equation*}
        R_S(\mathbf{w}^{*}_R)-R(\mathbf{w}^{*}_R)
        =O\left(\frac{\log(1/\delta)}{\sqrt{n}}+\sqrt{\frac{\theta\log(1/\delta)}{n}} \right)
    \end{equation*}
    with probability $1-\delta$.
   When $r(\mathbf{w}^{*}_R)=O(\sigma\|\mathbf{w}^{*}_R\|_2^2)$, $\sigma=O(n^{-\frac{1}{2}})$, and 
   $$\max\{\sup_{z}\left(f(\mathbf{w}^{*}_R;z)\right), \sup_{z,\tilde{z}}\left(g(\mathbf{w}^{*}_R;z,\tilde{z})\right)\}=O(\sqrt{n}),$$ 
   we have
    \begin{equation*}
        R(A(S))-R(\mathbf{w}^{*}_R)=O\left(n^{-\frac{1}{2}}\log_2n \log(1/\delta)\right)
    \end{equation*}
    with probability $1-\delta$ based on Theorem \ref{rrm} and the above decomposition of excess risk. 
\end{remark}

\begin{remark}
    We now apply Theorem \ref{rrm} to the pairwise constraint-guided sparse model \cite{liu2015pairwise}, which is inspired from the $\ell_1$-penalty and $\ell_{2,1}$-penalty used in Lasso \cite{tibshirani2011regression} and its variants \cite{zou2006adaptive,yuan2006model,simon2013sparse,friedman2010note}. The optimization objective of \cite{liu2015pairwise} can be formulated as 
      \begin{equation*}
     \frac1n\sum\limits_{i\in[n]} f\left(\mathbf{w}; z_{i}\right) +\frac{\lambda_1}{n(n-1)}\sum\limits_{i, j \in[n]: i \neq j} g\left(\mathbf{w} ; z_{i}, z_{j}\right)+\lambda_{2}\|\mathbf{w}\|_{1},
   \end{equation*}
   where $f\left(\mathbf{w}; z_{i}\right)$ is the general least square loss and the pairwise part is measured by 
   $$\sum_{\left({x}_{i}, {x}_{j}\right) \in \mathbf{M}}(\mathbf{w}^{T}{x}_{i}-\mathbf{w}^{T} {x}_{j})^{2}-\lambda_3\sum_{({x}_{i}, {x}_{j}) \in \mathbf{C}}(\mathbf{w}^{T} {x}_{i}-\mathbf{w}^{T} {x}_{j})^{2}.$$
   Here, $\mathbf{M}$ and $\mathbf{C}$ denote   the must-link set and the cannot-link set respectively,  and $\lambda_1, \lambda_2, \lambda_3$ are tuneable parameters. It is straightforward to verify that the above objective function is strongly-convex and Lipschitz. Therefore, our theoretical analysis provides the generalization bounds of the PPL model \cite{liu2015pairwise}. 
\end{remark}

\subsection{Optimistic generalization bounds}
This subsection further investigates the refined generalization bounds with the help of on-average loss stability in Definition \ref{Pointwise and pairwise On-average Stability} and on-average argument stability in Definition \ref{ppl argument}. The related proofs can be found in \emph{Supplementary Material B.4}. 

\begin{theorem}\label{connection}
If $A$ is PPL $\gamma$-on-average loss stable, then $$\mathbb{E}_{S}[R(A(S))-R_{S}(A(S))]\leq \gamma.$$
\end{theorem}

As illustrated in Table \ref{Comparison2}, this quantitative relation is consistent with the previous results for pointwise learning \cite{lei2020fine} and pairwise learning \cite{lei2021generalization}. A similar result for $\ell_1$ on-average argument stability is stated as follows.
\begin{corollary}\label{Thl1}
   Assume that $A$ is PPL $\ell_1$ $\gamma$-on-average argument stable. If the pointwise loss function $f(\mathbf{w};z)$ and the pairwise loss function $g(\mathbf{w};z,\tilde{z})$ are $L$-Lipschitz with respect to $\mathbf{w}$, then
   \begin{equation*}
       \mathbb{E}_{S,A}[R(A(S))-R_{S}(A(S))]\leq L\gamma.
   \end{equation*}
\end{corollary}

For completeness, we introduce the following result of $\ell_2$ on-average argument stability, which is a natural extension of  Theorem 2 part (b) \cite{lei2020fine} and  Theorem 1 \cite{lei2021generalization}, and removes the requirement on the $L$-Lipschitz condition of loss functions for PPL. 
\begin{theorem}\label{Thl2}
\cite{lei2020fine,lei2021generalization} Let $A$ be PPL $\ell_2$ $\gamma$-on-average argument stable and $\epsilon>0$. If  $f(\mathbf{w};z)$ and  $g(\mathbf{w};z,\tilde{z})$ are nonnegative and $\beta$-smooth with respect to $\mathbf{w}$, then
   \begin{align*}
       &\mathbb{E}_{S,A}[R(A(S))-R_{S}(A(S))]\\
       \leq& \frac{\beta}{\gamma}\left(\mathbb{E}_{S,A}\left[\tau R_S^{point}(A(S))+(1-\tau)R_S^{pair}(A(S))\right]\right)\\
       &+(\beta+\epsilon)\gamma\left(2-\frac{3}{2}\tau\right).
   \end{align*}
\end{theorem}

In the expectation viewpoint, the generalization error can be bounded by the empirical risk and the drift of model parameters induced by the changes of training data, which is illustrated in the following lemma. 
\begin{lemma}\label{smooth1}
    Assume that the pointwise loss function $f(\mathbf{w};z)$ and the pairwise loss function $g(\mathbf{w};z,\tilde{z})$ are $\beta$-smooth with respect to $\mathbf{w}$. Let $\epsilon>0$ and $\tau\in[0,1]$. Then,
    \begin{align*}
     &\mathbb{E}\left[R(A(S))-R_{S}(A(S))\right]\\ 
     \leq& \frac{\beta\tau \mathbb{E}\left[R_{S}^{point}(A(S))\right]}{\epsilon}+\frac{\beta(1-\tau) \mathbb{E}\left[R_{S}^{pair}(A(S))\right]}{\epsilon}\\
     &+\frac{(\epsilon+\beta)}{n}\left(2-\frac{3}{2}\tau\right) \sum_{i=1}^{n} \mathbb{E}\left[\left\|A\left(S_{i}\right)-A(S)\right\|_2^{2}\right]. 
    \end{align*}
\end{lemma}

\begin{theorem}
   Assume that the pointwise loss function $f(\mathbf{w};z)$ and the pairwise loss function $g(\mathbf{w};z,\tilde{z})$ are $\beta$-smooth with respect to the first argument. Suppose $A$ is defined by \eqref{as} and $\mathbf{w}^{*}$ is defined by \eqref{w*}. If $F_S$ is $\sigma$-strongly convex and $\beta \leq \sigma n / 4(2-\tau)$, then
   \begin{align*}
      &\mathbb{E}_{S}[F(A(S))]-F\left(\mathbf{w}^{*}\right)\nonumber\\
      \leq&\mathbb{E}_{S}[R(A(S))-R_{S}(A(S))]\nonumber\\
      \leq&\frac{\beta\tau \mathbb{E}\left[R_{S}^{point}(A(S))\right]}{\epsilon}+\frac{\beta(1-\tau) \mathbb{E}\left[R_{S}^{pair}(A(S))\right]}{\epsilon}\nonumber\\
      &+\frac{384\tau^2(\epsilon+\beta)\beta}{\sigma^2n^2}\left(2-\frac{3}{2}\tau\right)\mathbb{E}\left[R_{S}^{point}(A(S))\right]\nonumber\\
      &+\frac{768(1-\tau)^2(\epsilon+\beta)\beta}{\sigma^2n^2}\left(2-\frac{3}{2}\tau\right)\mathbb{E}\left[R_{S}^{pair}(A(S))\right]\label{opt2}.
   \end{align*}
\end{theorem}

   \begin{remark}
       If $r(\mathbf{w})=O\left(\sigma\|\mathbf{w}\|^2_2\right)$, we can get
   \begin{align*}
        &\mathbb{E}_{S}[R(A(S))]-R(\mathbf{w}^{*}_{R})\\
        =&O\left(\frac{R^{point}(\mathbf{w}^{*}_{R})+R^{pair}(\mathbf{w}^{*}_{R})}{n\sigma}\right)+O\left((n^{-1}+\sigma)\|\mathbf{w}^{*}_{R}\|_2^2\right),
   \end{align*}
      where $\mathbf{w}^{*}_{R}$ is defined by \eqref{wr*}.
   Furthermore, taking
   $$
   \sigma=\max\left\{\frac{12\beta}{n}, \sqrt{\frac{R^{point}(\mathbf{w}^{*}_{R})+R^{pair}(\mathbf{w}^{*}_{R})}{n\|\mathbf{w}^{*}_{R}\|_2^2}}\right\},
   $$  
   we can conclude that
    \begin{align*}
        &\mathbb{E}_{S}[R(A(S))]-R(\mathbf{w}^{*}_{R})\\
        =&O\left(\frac{\|\mathbf{w}^{*}_{R}\|_2}{\sqrt{n}}\left[\sqrt{R^{point}(\mathbf{w}^{*}_{R})+R^{pair}(\mathbf{w}^{*}_{R})}\right]+\frac{\|\mathbf{w}^{*}_{R}\|_2^2}{n}\right).
   \end{align*}
   Moreover, when $$\max\{R^{point}(\mathbf{w}^{*}_{R}), R^{pair}(  \mathbf{w}^{*}_{R})\}=O\left( \frac{\|\mathbf{w}^{*}_{R}\|_2^2}{n}\right),$$ we get the fast convergence rate
   \begin{equation*}
       \mathbb{E}_{S}[R(A(S))]-R(\mathbf{w}^{*}_{R})=O\left(\frac{\|\mathbf{w}^{*}_{R}\|_2^2}{n} \right).
   \end{equation*}
   The derived rate $O(n^{-1})$ often is considered as tightness enough in statistical learning theory \cite{shalev2010learnability,hardt2016train}. 
   \end{remark}
   
\begin{remark}
 It should be noticed that the current result is consistent with the pointwise setting \cite{lei2020fine} as $\tau=1$, with the pairiwise setting \cite{lei2020sharper} as $\tau=0$. Our convergence analysis of PPL setting  covers more complicated learning algorithms (e.g., algorithms  described in Table \ref{Comparison1}) due to the flexibility of $\tau\in[0,1]$.
\end{remark}

\section{Conclusion}
This paper focuses on establishing the generalization bounds of PPL by means of  algorithmic stability analysis. After characterizing the quantitative relationship between generalization error and algorithmic stability, we establish the upper bounds of excess risk of SGD and RRM for PPL. Our stability-based analysis fills the gap of statistical learning theory in part for  the related PPL algorithms. In the future, it is interesting to further investigate the stability-based generalization of SGD for PPL under non-i.i.d sampling, e.g., Markov chain sampling \cite{sun-markov-18,wang-markov-2022}.

\section{Acknowledgments}
This work was supported in part by National
Natural Science Foundation of China under Grant No. 12071166 and by the Fundamental Research Funds for the Central
Universities of China under Grant 2662020LXQD002. We are grateful to the anonymous AAAI reviewers for their constructive comments.
\bibliography{ref}{}

\onecolumn

\section{A.~~ Notations}
The main notations of this paper are summarized in Table \ref{Notations}.
\begin{table}[h]
    \centering
	\renewcommand\arraystretch{1.2}
	\begin{tabular}{|l|l|l|l|}
		\hline
        $\mathcal{Z}$& Sample space&$S$& Training dataset $\left\{z_{1}, \ldots, z_{n}\right\}$\\
        \hline
        $z_i$&$i$-th training example&$\rho$&Probability measure\\
        \hline
        $\mathcal{W}$&Model parameter space&$\mathbf{w}$&Model parameter\\
        \hline
        $f(\mathbf{w} ; z)$ & Pointwise loss function&$g(\mathbf{w} ; z, \tilde{z})$&Pairwise loss function\\
        \hline
        $\ell(\mathbf{w} ; z, \tilde{z})$ & PPL loss function&SGD & Stochastic gradient descent\\
        \hline
        ERM & Empirical risk minimization&RRM & Regularized risk minimization\\
        \hline
        $A$& Given algorithm&$A(S)$& Output model\\
        \hline
        $R^{point}$&Pointwise expected risk&$R^{point}_S$ &Pointwise empirical risk\\
        \hline
        $R^{pair}$&Pairwise expected risk&$R^{pair}_S$&Pairwise empirical risk\\
        \hline
        $R(\mathbf{w})$&$\tau R^{point}(\mathbf{w})+(1-\tau) R^{pair}(\mathbf{w})$&$R_S(\mathbf{w})$&$\tau R^{point}_S(\mathbf{w})+(1-\tau) R^{pair}_S(\mathbf{w})$ \\ 
        \hline
        $F_S(\mathbf{w})$&$R_S(\mathbf{w})+r(\mathbf{w})$&$\tau$&PPL tuning parameter\\
        \hline
        $Gen$& $R(A(S))-R_{S}(A(S))$&$\mathbb{E}_{Gen}$& $\mathbb{E}_{S,A}[R(A(S))-R_{S}(A(S))]$\\
        \hline
        $\mathbf{w}^{*}$&$\arg \min _{\mathbf{w} \in \mathcal{W}}[R(\mathbf{w})+r(\mathbf{w})]$&$\mathbf{w}^{*}_{R}$&$\arg \min _{\mathbf{w} \in \mathcal{W}}[R(\mathbf{w})]$\\
        \hline
        $\gamma$& Stability parameter&$L$& Lipschitz parameter\\
        \hline
        $\beta$& Smoothness parameter&$\sigma$& Strong convexity parameter\\
        \hline
        $U_{point}$&Pointwise uniform stability&$U_{pair}$&Pairwise uniform stability\\
        \hline
        $V_{point}$&Pointwise on-average loss stability&$V_{pair}$&Pairwise on-average loss stability\\
        \hline
        $H_{point}$&$\frac{1}{n}\sum_{i=1}^{n}\|A(S)-A(S_i)\|_2$&$H_{pair}$&$\frac{\sum_{i,j\in[n]:i\neq j}\|A(S)-A(S_{i,j})\|_2}{n(n-1)}$\\
		\hline
		$n$&Sample size&$d$&the dimension of $\mathcal{X}$\\
		\hline
		$T$&Iteration number&$\eta_t$&Step size\\
		\hline
		$\nabla$&the gradient of loss function&$\mathbb{E}_{S}[\cdot]$&  Conditional expectation about $S$\\
		\hline
		$\|Y\|_p$&$(\mathbb{E}|Y|^p)^{\frac{1}{p}}$&$[n]$&$\{1, \ldots, n\}$\\
		\hline
	\end{tabular}
	\caption{Table of Notations.}
	\label{Notations}
\end{table}

\section{B. Proofs of main theoretical results}
\begin{figure}[h]
	\centering
	\includegraphics[width=0.9\linewidth]{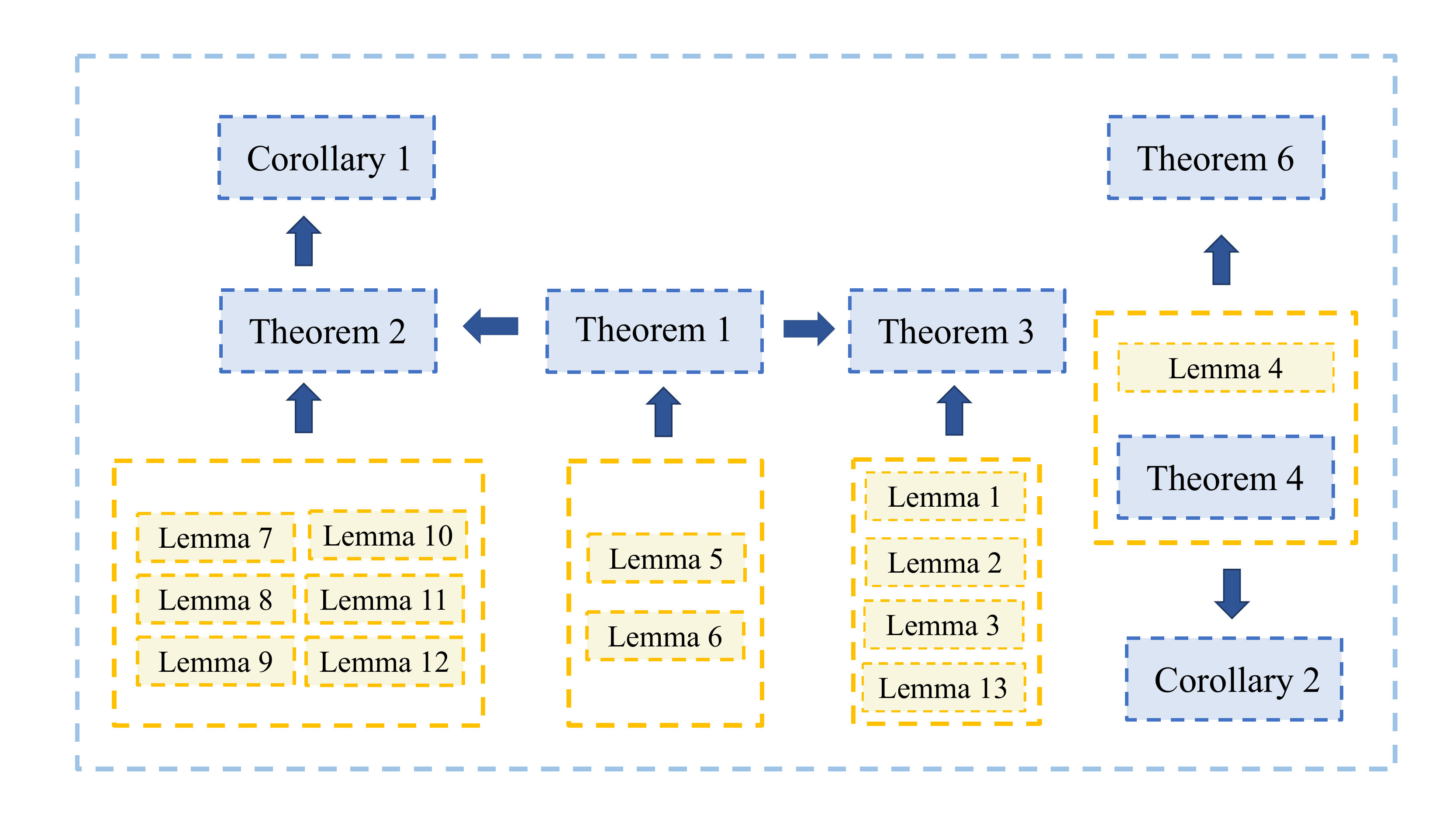}
	\caption{Box of theoretical results for Pointwise and Pairwise Learning.}
	\label{fig1}
\end{figure}

To improve the readability of theoretical analysis, we state the outlines for the proofs of Theorems 1-6 in Fig \ref{fig1}
\subsection{B.1~~~ Proof of Theorem 1}

To prove Theorem 1, we need to introduce some lemmas as follows.

\begin{lemma}\label{lemma 1}
	\cite{bousquet2020sharper}
	Let $S=\left\{z_{1}, \ldots, z_{n}\right\}$ be a set of independent random variables each taking values in $\mathcal{Z}$ and let $d_{1}, \ldots, d_{n}$ be some functions $d_{i}: \mathcal{Z}^{n} \mapsto \mathbb{R}$ such that the following holds for any $i \in[n]$
	\begin{itemize}
		\item $\left|\mathbb{E}_{S \backslash\left\{z_{i}\right\}}\left[d_{i}(S)\right]\right| \leq M$ almost surely (a.s.), where $S \backslash\left\{z_{i}\right\}=\{z_{1}, \ldots,z_{i-1},z_{i+1},\ldots, z_{n}\}$,
		\item $\mathbb{E}_{z_{i}}\left[d_{i}(S)\right]=0$ a.s.,
		\item $g_i$ has a bounded difference $\alpha$ with respect to all variables except the i-th variable
		$$
		\left|d_{i}(z_{1}, \ldots, z_{n})-d_{i}\left(z_{1}, \ldots, z_{j-1}, z_{j}^{\prime \prime}, z_{j+1}, \ldots, z_{n}\right)\right| \leq \alpha,
		$$
		where $j \in[n]$ with $j \neq i$, and $z_{j}^{\prime \prime} \in \mathcal{Z}$.
	\end{itemize}
	Then, for any $p \geq 2$
		$$
		\left\|\sum_{i=1}^{n} d_{i}(S)\right\|_{p} \leq 12 \sqrt{2} p n \alpha\left\lceil\log _{2} n\right\rceil+4 M \sqrt{p n} .
		$$
\end{lemma}

\begin{lemma}\label{lemma 2}
    \cite{bousquet2020sharper}
	Let $a, b\geq 0$ and let $Y$ be a random variable with $\|Y\|_{p} \leq \sqrt{p} a+p b$ for any $p \geq 2$. Then for any $\delta \in(0,1)$ with probability at least $1-\delta$
	$$
	|Y| \leq e\left(a \sqrt{\log \left(\frac{e}{\delta}\right)}+b\log \left(\frac{e}{\delta}\right)\right) .
	$$
\end{lemma}

\subsubsection{Proof of Theorem 1}
From $(2)$, we know the generalization error 
\begin{align*}
    &R(A(S))-R_{S}(A(S))\\
   = &\tau R^{point}(A(S))+(1-\tau)R^{pair}(A(S))-\tau R_{S}^{point}(A(S))-(1-\tau)R_{S}^{pair}(A(S))\\
    =&\tau\mathbb{E}_{z}[f(\mathbf{w} ; z)]+(1-\tau)\mathbb{E}_{z, \tilde{z}}[g(\mathbf{w} ; z, \tilde{z})]-\frac{\tau}{n} \sum_{i\in[n]} f\left(\mathbf{w} ; z_{i}\right)-\frac{(1-\tau)}{n(n-1)} \sum_{i, j \in[n]: i \neq j} g\left(\mathbf{w} ; z_{i}, z_{j}\right).
\end{align*}
It is easy to see that the above generalization gap can be divided into two parts associated with pointwise loss and pairwise loss, respectively.
To bound the pointwise part, we consider
\begin{align*}
	&n \mathbb{E}_{Z}[f(A(S) ; Z)]-\sum_{i=1}^n f\left(A(S) ; z_{i}\right)\\
	=&\sum_{i=1}^n \mathbb{E}_{Z}\left[f(A(S) ; Z)-\mathbb{E}_{z_{i}^{\prime}} \left[f\left(A\left(S_{i}\right) ; Z\right)\right]\right] 
	+\sum_{i=1}^n \mathbb{E}_{z_{i}^{\prime} }\left[\mathbb{E}_{Z }\left[f\left(A\left(S_{i}\right) ; Z\right)\right]-f\left(A\left(S_{i}\right) ; z_{i}\right)\right]\\
	&+\sum_{i=1}^n \mathbb{E}_{z_{i}^{\prime} }\left[f\left(A\left(S_{i}\right) ; z_{i}\right)-f\left(A(S) ; z_{i}\right)\right],
\end{align*}
where $S_i$ is defined by $(3)$.
According to the definition of PPL uniform stability in Definition $1$, we know
\begin{equation*}
    \left|f(A(S) ; Z)-\mathbb{E}_{z_{i}^{\prime}} \left[f\left(A\left(S_{i}\right) ; Z\right)\right]\right| \leq  \gamma, \quad\left|f\left(A\left(S_{i}\right) ; z_{i}\right)-f\left(A(S) ; z_{i}\right)\right| \leq  \gamma.
\end{equation*}
Therefore,
\begin{equation}\label{pointwise gap}
	\left|n \mathbb{E}_{Z}[f(A(S) ; Z)]-\sum_{i=1} f\left(A(S) ; z_{i}\right)\right| \leq 2 n \gamma+\left|\sum_{i=1}^{n} d_{i}(S)\right|,
\end{equation}
where 
$$
d_{i}(S)=\mathbb{E}_{z_{i}^{\prime}}\left[\mathbb{E}_{Z}\left[f\left(A\left(S_{i}\right) ; Z\right)\right]\right]-\mathbb{E}_{z_{i}^{\prime}}\left[f\left(A\left(S_{i}\right) ; z_{i}\right)\right]=d_{i}^{(1)}(S_i)-d_{i}^{(2)}(S_i), \quad \forall i \in[n] .
$$
Due to the boundedness assumption of loss function, we know
$$
\left|\mathbb{E}_{S \backslash\left\{z_{i}\right\}}\left[d_{i}(S)\right]\right| \leq 2 M, \quad \forall i \in[n] .
$$
Since $z_{i}$ is independent of $S_{i}$, there holds
$$
\mathbb{E}_{z_{i}}\left[\mathbb{E}_{Z}\left[f\left(A\left(S_{i}\right) ; Z\right)\right]-\left[f\left(A\left(S_{i}\right) ; z_{i}\right)\right]\right]=0,
$$
which means $\mathbb{E}_{z_{i}}\left[d_{i}\right]=0$. 

By using the PPL uniform stability, we derive that
\begin{equation*}
    \left|d_{i}^{(1)}(S_i)-d_{i}^{(1)}(S_i^{(m)})\right|=\left|\mathbb{E}_{z_{i}^{\prime}} \mathbb{E}_{Z}\left[f\left(A\left(S_{i}\right) ; Z\right)\right]-\mathbb{E}_{z_{i}^{\prime}} \mathbb{E}_{Z}\left[f\left(A\left(S_{i}^{(m)}\right) ; Z\right)\right]\right| \leq \gamma,
\end{equation*}
where $S_{i}^{(m)}$ is the set $S_{i}$ replacing the $m$-th element with $z_{m}^{\prime}$.
Similarly, we can also conduct
\begin{equation*}
    \left|d_{i}^{(2)}(S_i)-d_{i}^{(2)}(S_i^{(m)})\right|=\left|\mathbb{E}_{z_{i}^{\prime}} \left[f\left(A\left(S_{i}\right) ; z_{i}\right)\right]-\mathbb{E}_{z_{i}^{\prime}} \left[f\left(A\left(S_{i}^{(m)}\right) ; z_{i}\right)\right]\right| \leq \gamma.
\end{equation*}
Therefore, applying Lemma \ref{lemma 1} with $\alpha=2 \gamma$, we have
$$
\left\|\sum_{i=1}^n d_{i}\right\|_{p} \leq 24 \sqrt{2} pn \gamma\left\lceil\log _{2}n\right\rceil+4 M \sqrt{pn}.
$$
According to Lemma \ref{lemma 2}, we  derive that 
$$
\left|\sum_{i=1}^{n} d_{i}\right| \leq e\left(4 M \sqrt{n}\sqrt{\log (e / \delta)}+24\sqrt{2} n \gamma\left\lceil\log _{2}n\right\rceil \log (e / \delta)\right)$$
with probability at least $1-\delta$. Furthermore, we plug the above inequality back into \eqref{pointwise gap} to derive 
\begin{equation*}
    \left|R_{S}^{point}(A(S))-R^{point}(A(S))\right|\leq 2\gamma+e\left(4 Mn^{-\frac{1}{2}} \sqrt{\log (e / \delta)}+24 \sqrt{2} \gamma\left\lceil\log _{2}(n)\right\rceil \log (e / \delta)\right)
\end{equation*}
 with probability at least $1-\delta$.
 
Theorem 1 of \citet{lei2020sharper} assures that the pairwise part satisfies 
\begin{equation*}
	\left|R_{S}^{pair}(A(S))-R^{pair}(A(S))\right| \leq 4\gamma+e\left(16 M(n-1)^{-\frac{1}{2}} \sqrt{\log (e / \delta)}+48 \sqrt{2} \gamma\left\lceil\log _{2}(n-1)\right\rceil \log (e / \delta)\right)
\end{equation*}
with probability at least $1-\delta$.
The desired result follows by combining the two estimations for pointwise and pairwise parts.
\\
\qed

\subsection{B.2~~~ Proof of Theorem 2}
To prove Theorem 2, we introduce the PPL non-expansiveness operator which is motivated by \citet{hardt2016train} and extended in \citet{lei2020sharper}. We also need to introduce some properties of the PPL loss function when the pointwise loss function part and the pairwise loss function part change.
\begin{lemma}\cite{hardt2016train}
	Assume the pointwise loss function $f\left(\mathbf{w} ; z\right)$ is convex and $\beta$-smooth with respect to $\mathbf{w}$ for all $z \in \mathcal{Z}$. Then for all $\eta \leq 2 / \beta$ there holds
	$$
	\left\|\mathbf{w}-\eta f^{\prime}\left(\mathbf{w} ; z\right)-\mathbf{w}^{\prime}+\eta f^{\prime}\left(\mathbf{w}^{\prime} ; z\right)\right\|_{2} \leq\left\|\mathbf{w}-\mathbf{w}^{\prime}\right\|_{2} .
	$$
\end{lemma}
Correspondingly, the non-expansiveness operators of the pairwise and the pointwise are the same except for the functional forms.
\begin{lemma}
	Assume the pairwise loss function $g\left(\mathbf{w} ; z, \tilde{z}\right)$ is convex and $\beta$-smooth with respect to $\mathbf{w}$ for all $z, \tilde{z}\in \mathcal{Z}$. Then for all $\eta \leq 2 / \beta$ there holds
	$$
	\left\|\mathbf{w}-\eta g^{\prime}\left(\mathbf{w} ; z, \tilde{z}\right)-\mathbf{w}^{\prime}+\eta g^{\prime}\left(\mathbf{w}^{\prime} ; z, \tilde{z}\right)\right\|_{2} \leq\left\|\mathbf{w}-\mathbf{w}^{\prime}\right\|_{2} .
	$$
\end{lemma}

\begin{lemma}\label{convex}
	If the pointwise loss function $f\left(\mathbf{w} ; z\right)$ and the pairwise loss function $g\left(\mathbf{w} ; z, \tilde{z}\right)$ are convex with respect to $\mathbf{w}$ for $z, \tilde{z} \in \mathcal{Z}$, then the PPL loss function $\ell\left(\mathbf{w} ; z, \tilde{z}\right)$ is convex with respect to $\mathbf{w}$.
\end{lemma}
\subsubsection{Proof of Lemma \ref{convex}}
According to the definition of convex function, we can easily get the above conclusion, and the proof process is omitted here.\\
\qed
\begin{lemma}\label{smooth}\label{lips}
	If the pointwise loss function $f\left(\mathbf{w} ; z\right)$ and the pairwise loss function $g\left(\mathbf{w} ; z, \tilde{z}\right)$ are $\beta$-smooth with respect to $\mathbf{w}$ for $z, \tilde{z} \in \mathcal{Z}$, then the PPL loss function $\ell\left(\mathbf{w} ; z, \tilde{z}\right)$ is $\beta$-smooth with respect to $\mathbf{w}$. Similarly, if the pointwise loss function and the pairwise loss function are $L$-Lipschitz with respect to $\mathbf{w}$ for $z, \tilde{z} \in \mathcal{Z}$, then the PPL loss function is $L$-Lipschitz with respect to $\mathbf{w}$.
\end{lemma}

\subsubsection{Proof of Lemma \ref{smooth}}
Based on the definitions of smoothness, the following inequalities hold for $f$ and $g$:
$$
\tau\|\nabla f(u)-\nabla f(v)\|_2 \leq \tau\beta\|u-v\|_2,
$$
$$
(1-\tau)\|\nabla g(u)-\nabla g(v)\|_2 \leq (1-\tau)\beta\|u-v\|_2,
$$
$$
\tau\|\nabla f(u)-\nabla f(v)\|_2+(1-\tau)\|\nabla g(u)-\nabla g(v)\|_2 \leq \beta\|u-v\|_2.
$$
From the sub-additivity of $\|\cdot\|_2$, we know 
$$
\|\tau\nabla f(u)+(1-\tau)\nabla g(u)-\tau\nabla f(v)-(1-\tau)\nabla g(v)\|_2 \leq \beta\|u-v\|_2,
$$
i.e.,
$$
\|\nabla \ell(u)-\nabla \ell(v)\|_2 \leq \beta\|u-v\|_2.
$$
The proof of the Lipschitz continuity is similar to the above process, thus we omit it here.\\
The proof is completed.\\
\qed

\par Since the PPL loss function is convex and $\beta$-smooth, We can also get the non-expansiveness of the PPL as follows.
\begin{lemma}\label{non}
	Assume for all $z, \tilde{z}\in \mathcal{Z}$, the functions $\mathbf{w} \mapsto f\left(\mathbf{w} ; z\right)$ and $\mathbf{w} \mapsto g\left(\mathbf{w} ; z, \tilde{z}\right)$ are both convex and $\beta$-smooth. Then for all $\eta \leq 2 / \beta$ there holds
	$$
	\left\|\mathbf{w}-\eta \ell^{\prime}\left(\mathbf{w} ; z, \tilde{z}\right)-\mathbf{w}^{\prime}+\eta \ell^{\prime}\left(\mathbf{w}^{\prime} ; z, \tilde{z}\right)\right\|_{2} \leq\left\|\mathbf{w}-\mathbf{w}^{\prime}\right\|_{2} .
	$$
\end{lemma}
In addition, we also need to introduce a concentration inequality \cite{boucheron2013concentration,shalev2014understanding} which is useful for developing high-probability bounds in the following part.
\begin{lemma}\label{bounds}
	(Chernoff's Bound).Let $X=\sum_{t=1}^{T} X_{t}$ where $X_{1}, \ldots, X_{T}$ is independent random variables taking values in $\{0,1\}$ and $\mu=\mathbb{E}[X]$. Then for any $\delta_1$, we have $X \leq (1+\delta_1)\mu$ with probability at least $1-exp(-\mu\delta_1^2/(3+\delta_1))$. Moreover, for any $\delta \in(0,1)$, we have
	$$
	X \leq \mu+\log(1/\delta)+\sqrt{2\mu\log(1/\delta)}
	$$
	with probability at least $1-\delta$.
\end{lemma}
\subsubsection{Proof of Lemma \ref{bounds}}

We now establish the uniform stability of PPL SGD.

\subsubsection{Proof of Theorem 2}
Let $\left\{\mathbf{w}_{t}\right\},\left\{\mathbf{w}_{t}^{\prime}\right\}$ be produced by PPL SGD on $S$ and $S^{\prime}$, then
$$
\mathbf{w}_{t+1}=\mathbf{w}_{t}-\eta_{t}\left(\tau f^{\prime}\left(\mathbf{w}_{t} ; z_{i_{t}}\right)+(1-\tau)g^{\prime}\left(\mathbf{w}_{t} ; z_{i_{t}}, z_{j_{t}}\right)\right)
$$
and
$$
\mathbf{w}_{t+1}^{\prime}=\mathbf{w}_{t}^{\prime}-\eta_{t}\left(\tau f^{\prime}\left(\mathbf{w}_{t}^{\prime} ; z_{i_{t}}^{\prime}\right)+(1-\tau)g^{\prime}\left(\mathbf{w}_{t}^{\prime} ; z_{i_{t}}^{\prime}, z_{j_{t}}^{\prime}\right)\right).
$$

In the sequel, let us consider three cases. Firstly, if $i_{t} \in[n-1]$ and $j_{t} \in[n-1]$, $i_t \neq j_t$, then
\begin{align*}
	&\mathbf{w}_{t+1}-\mathbf{w}_{t+1}^{\prime}\\ =&\mathbf{w}_{t}-\eta_{t} \left(\tau f^{\prime}\left(\mathbf{w}_{t} ; z_{i_{t}}\right)+(1-\tau)g^{\prime}\left(\mathbf{w}_{t} ; z_{i_{t}}, z_{j_{t}}\right)\right)-\mathbf{w}_{t}^{\prime}+\eta_{t}\left(\tau f^{\prime}\left(\mathbf{w}_{t}^{\prime} ; z_{i_{t}}^{\prime}\right)+(1-\tau)g^{\prime}\left(\mathbf{w}_{t}^{\prime} ; z_{i_{t}}^{\prime}, z_{j_{t}}^{\prime}\right)\right) \\
    =&\mathbf{w}_{t}-\eta_{t} \left(\tau f^{\prime}\left(\mathbf{w}_{t} ; z_{i_{t}}\right)+(1-\tau)g^{\prime}\left(\mathbf{w}_{t} ; z_{i_{t}}, z_{j_{t}}\right)\right)-\mathbf{w}_{t}^{\prime}+\eta_{t}\left(\tau f^{\prime}\left(\mathbf{w}_{t}^{\prime} ; z_{i_{t}}\right)+(1-\tau)g^{\prime}\left(\mathbf{w}_{t}^{\prime} ; z_{i_{t}}, z_{j_{t}}\right)\right).
\end{align*}

It then follows from Lemma \ref{non} that
$$
\left\|\mathbf{w}_{t+1}-\mathbf{w}_{t+1}^{\prime}\right\|_{2} \leq\left\|\mathbf{w}_{t}-\mathbf{w}_{t}^{\prime}\right\|_{2} \text {. }
$$
Secondly, if $i_{t} = n$ and $j_{t} \in[n-1]$, then
\begin{align*}
	&\left\|\mathbf{w}_{t+1}-\mathbf{w}_{t+1}^{\prime}\right\|_{2}\\ =&\left\|\mathbf{w}_{t}-\eta_{t} \left(\tau f^{\prime}\left(\mathbf{w}_{t} ; z_{i_{t}}\right)+(1-\tau)g^{\prime}\left(\mathbf{w}_{t} ; z_{i_{t}}, z_{j_{t}}\right)\right)-\mathbf{w}_{t}^{\prime}+\eta_{t}\left(\tau f^{\prime}\left(\mathbf{w}_{t}^{\prime} ; z_{i_{t}}^{\prime}\right)+(1-\tau)g^{\prime}\left(\mathbf{w}_{t}^{\prime} ; z_{i_{t}}^{\prime}, z_{j_{t}}^{\prime}\right)\right)\right\|_{2} \\
	\leq& \left\|\mathbf{w}_{t}-\mathbf{w}_{t}^{\prime}\right\|_{2}+\tau\left\|\eta_{t} f^{\prime}\left(\mathbf{w}_{t}^{\prime} ; z_{i_{t}}^{\prime} \right)-\eta_{t} f^{\prime}\left(\mathbf{w}_{t} ; z_{i_{t}}\right)\right\|_{2}+(1-\tau)\left\|\eta_{t} g^{\prime}\left(\mathbf{w}_{t}^{\prime} ; z_{i_{t}}^{\prime}, z_{j_{t}}^{\prime}\right)-\eta_{t} g^{\prime}\left(\mathbf{w}_{t} ; z_{i_{t}}, z_{j_{t}}\right)\right\|_{2} \\
	\leq& \left\|\mathbf{w}_{t}-\mathbf{w}_{t}^{\prime}\right\|_{2}+2 \tau\eta_{t} L+2 (1-\tau)\eta_{t} L\\
	=&\left\|\mathbf{w}_{t}-\mathbf{w}_{t}^{\prime}\right\|_{2}+2\eta_{t} L,
\end{align*}

where the last inequality is due to the L-Lipschitz continuity of the loss functions $f$ and $g$.\\
Finally, if $i_{t} \in[n-1]$ and $j_{t} = n$, then
\begin{align*}
	&\left\|\mathbf{w}_{t+1}-\mathbf{w}_{t+1}^{\prime}\right\|_{2}\\ =&\left\|\mathbf{w}_{t}-\eta_{t} \left(\tau f^{\prime}\left(\mathbf{w}_{t} ; z_{i_{t}}\right)+(1-\tau)g^{\prime}\left(\mathbf{w}_{t} ; z_{i_{t}}, z_{j_{t}}\right)\right)-\mathbf{w}_{t}^{\prime}+\eta_{t}\left(\tau f^{\prime}\left(\mathbf{w}_{t}^{\prime} ; z_{i_{t}}^{\prime}\right)+(1-\tau)g^{\prime}\left(\mathbf{w}_{t}^{\prime} ; z_{i_{t}}^{\prime}, z_{j_{t}}^{\prime}\right)\right)\right\|_{2} \\
	\leq& \left\|\mathbf{w}_{t}-\eta_{t} \tau f^{\prime}\left(\mathbf{w}_{t} ; z_{i_{t}} \right)-\mathbf{w}_{t}^{\prime}+\eta_{t} \tau f^{\prime}\left(\mathbf{w}_{t}^{\prime}; z_{i_{t}}\right)\right\|_{2}+(1-\tau)\left\|\eta_{t} g^{\prime}\left(\mathbf{w}_{t}^{\prime} ; z_{i_{t}}^{\prime}, z_{j_{t}}^{\prime}\right)-\eta_{t} g^{\prime}\left(\mathbf{w}_{t} ; z_{i_{t}}, z_{j_{t}}\right)\right\|_{2} \\
	\leq& \left\|\mathbf{w}_{t}-\mathbf{w}_{t}^{\prime}\right\|_{2}+2 \eta_{t} L(1-\tau).
\end{align*}
Combining the above three cases, we derive
$$
\left\|\mathbf{w}_{t+1}-\mathbf{w}_{t+1}^{\prime}\right\|_{2} \leq\left\|\mathbf{w}_{t}-\mathbf{w}_{t}^{\prime}\right\|_{2}+2 \eta_{t} L \mathbb{I}\left[i_{t}=n\right]+2 \eta_{t} L(1-\tau)\mathbb{I}\left[j_{t}=n\right],
$$
where $\mathbb{I}[\cdot]$ is the indicator function. Iterating over the above formula ($w_{1}=w_{1}^{\prime}$), there holds 
\begin{equation}\label{iter}
    \left\|\mathbf{w}_{t+1}-\mathbf{w}_{t+1}^{\prime}\right\|_{2} \leq 2 L \sum_{k=1}^{t} \eta_{k} \mathbb{I}\left[i_{k}=n\right]+2 L (1-\tau)\sum_{k=1}^{t} \eta_{k} \mathbb{I}\left[j_{k}=n\right].
\end{equation}
With the Lipschitz continuity, for all $z, \tilde{z} \in \mathcal{Z}$, we know
$$
\left|\ell\left(\mathbf{w}_{t+1} ; z, \tilde{z}\right)-\ell\left(\mathbf{w}_{t+1}^{\prime} ; z, \tilde{z}\right)\right| \leq L\left\|\mathbf{w}_{t+1}-\mathbf{w}_{t+1}^{\prime}\right\|_{2},
$$
hence
$$
\left|\ell\left(\mathbf{w}_{t+1} ; z, \tilde{z}\right)-\ell\left(\mathbf{w}_{t+1}^{\prime} ; z, \tilde{z}\right)\right| \leq 2 L^2 \sum_{k=1}^{t} \eta_{k} \mathbb{I}\left[i_{k}=n\right]+2 L^2(1-\tau) \sum_{k=1}^{t} \eta_{k} \mathbb{I}\left[j_{k}=n\right].
$$

We assume $X_k=\mathbb{I}[i_k=n]$, it is easy to know that
$$
\mathbb{E}[X_k]=Pr\{i_k=n\}=\frac{1}{n}.
$$
Applying Lemma \ref{bounds} with $X_{k}=\mathbb{I}\left[i_{k}=n\right]$, we obtain with probability $1-\delta$,
$$
\sum_{k=1}^{t} X_{k} \leq \mu+\log(1/\delta)+\sqrt{2\mu\log(1/\delta)},
$$
where $\mu=\sum_{k=1}^{t} \mathbb{E}\left[X_{t}\right] \leq t / n$. It then follows with probability $1-\delta$ that
$$
\sum_{k=1}^{t} X_{k} \leq \frac{t}{n}+\log(1/\delta)+\sqrt{2 n^{-1} t \log (1 / \delta)} .
$$
We can also derive the same results for $Y_{k}=\mathbb{I}\left[j_{k}=n\right]$
$$
\sum_{k=1}^{t} Y_{k} \leq \frac{t}{n}+\log(1/\delta)+\sqrt{2 n^{-1} t \log (1 / \delta)} .
$$
Combining the above two inequalities with \eqref{iter}, we derive the following inequality with probability $1-\delta$
$$
\left\|\mathbf{w}_{t+1}-\mathbf{w}_{t+1}^{\prime}\right\|_{2} \leq 2L\eta(2-\tau)\left(\frac{t}{n}+\log(1/\delta)+\sqrt{2 n^{-1} t \log (1 / \delta)} \right).
$$
The proof is completed with $\eta_t=\eta$.\\
\qed

\subsection{B.3~~~ Proof of Theorem 3}

Next, we prove some important lemmas,
\begin{lemma}\label{lem}
	For any $S \in \mathcal{Z}^{n}$, define $A$ as $A(S)=\arg \min _{\mathbf{w} \in \mathcal{W}} F_{S}(\mathbf{w})$. For any $k \in[n]$, let $S_{k}$ be defined by $(3)$. Then
	\begin{align*}
	 	&F_{S}\left(A\left(S_{k}\right)\right)-F_{S}(A(S))\\
	 	\leq &\frac{\tau}{n}\left( f\left(A\left(S_{k}\right) ; z_{k}\right)- f\left(A\left(S_{k}\right) ; z_{k}^{\prime}\right)+f\left(A\left(S\right) ; z_{k}^{\prime}\right)-f\left(A\left(S\right) ; z_{k}\right)\right)\\
	 	&+\frac{1-\tau}{n(n-1)} \sum_{i \in[n]: i \neq k}\left[\left((g(A(S_{k}) ; z_{i}, z_{k})-g(A(S) ; z_{i}, z_{k}))+(g\left(A\left(S_{k}\right) ; z_{k}, z_{i}\right)-g\left(A(S) ; z_{k}, z_{i}\right))\right.\right. \\
	    &+\left.\left(g\left(A(S) ; z_{i}, z_{k}^{\prime}\right)-g\left(A\left(S_{k}\right) ; z_{i}, z_{k}^{\prime}\right)\right)+\left(g\left(A(S) ; z_{k}^{\prime}, z_{i}\right)-g\left(A\left(S_{k}\right) ; z_{k}^{\prime}, z_{i}\right)\right)\right].
	\end{align*}
\end{lemma}

\subsubsection{Proof of Lemma \ref{lem}}
Without loss of generality, we can assume $k=n$. 
Due to $A\left(S_{n}\right)$ is the minimizer of $F_{S_{n}}$, we know
\begin{align}
	&F_{S}\left(A\left(S_{n}\right)\right)-F_{S}(A(S)) \nonumber\\
	=&F_{S}\left(A\left(S_{n}\right)\right)-F_{S_{n}}\left(A\left(S_{n}\right)\right)+F_{S_{n}}\left(A\left(S_{n}\right)\right)-F_{S_{n}}(A(S))+F_{S_{n}}(A(S))-F_{S}(A(S)) \nonumber\\
	\leq& F_{S}\left(A\left(S_{n}\right)\right)-F_{S_{n}}\left(A\left(S_{n}\right)\right)+F_{S_{n}}(A(S))-F_{S}(A(S)) . \label{eq}
\end{align}
By the definitions of $F_{S}$ and $F_{S_{n}}$, we obtain 
\begin{align}
	&F_{S}\left(A\left(S_{n}\right)\right)-F_{S_{n}}\left(A\left(S_{n}\right)\right)\nonumber\\
	=&\frac{\tau}{n}\left(\sum_{i=1}^{n} f^{\prime}\left(A\left(S_{n}\right) ; z_{i}\right)-\sum_{i=1}^{n-1} f^{\prime}\left(A\left(S_{n}\right) ; z_{i}\right)- f^{\prime}\left(A\left(S_{n}\right) ; z_{n}^{\prime}\right)\right)+\frac{1-\tau}{n(n-1)}\sum_{i, j \in[n]: i \neq j} g\left(A\left(S_{n}\right) ; z_{i}, z_{j}\right) \nonumber\\
    &-\frac{1-\tau}{n(n-1)}\left(\sum_{i, j \in[n-1]: i \neq j} g\left(A\left(S_{n}\right) ; z_{i}, z_{j}\right)+\sum_{i \in[n-1]} g\left(A\left(S_{n}\right) ; z_{i}, z_{n}^{\prime}\right)+\sum_{i \in[n-1]} g\left(A\left(S_{n}\right) ; z_{n}^{\prime}, z_{i}\right)\right) \nonumber\\
    =&\frac{\tau}{n}\left(f^{\prime}\left(A\left(S_{n}\right) ; z_{n}\right)- f^{\prime}\left(A\left(S_{n}\right) ; z_{n}^{\prime}\right)\right)\nonumber\\
    &+\frac{(1-\tau)}{n(n-1)}\sum_{i \in[n-1]}\left(g\left(A\left(S_{n}\right) ; z_{i}, z_{n}\right)+g\left(A\left(S_{n}\right) ; z_{n}, z_{i}\right)-g\left(A\left(S_{n}\right) ; z_{i}, z_{n}^{\prime}\right)-g\left(A\left(S_{n}\right) ; z_{n}^{\prime}, z_{i}\right)\right).\label{f1}
\end{align}
Similarly, we know
\begin{align}
    F_{S_{n}}\left(A\left(S\right)\right)-F_{S}\left(A\left(S\right)\right)
    = &\frac{\tau}{n}\left(f^{\prime}\left(A\left(S\right) ; z_{n}^{\prime}\right)-f^{\prime}\left(A\left(S\right) ; z_{n}\right)\right)
    +\frac{(1-\tau)}{n(n-1)}\sum_{i \in[n-1]}\left(g\left(A(S) ; z_{i}, z_{n}^{\prime}\right)\right.\nonumber\\
    &+\left.g\left(A(S) ; z_{n}^{\prime}, z_{i}\right)-g\left(A(S) ; z_{i}, z_{n}\right)-g\left(A(S) ; z_{n}, z_{i}\right)\right).\label{f2} 
\end{align}
We can combine \eqref{f1} and \eqref{f2} to derive
\begin{align*}
	&\left(F_{S}\left(A\left(S_{n}\right)\right)-F_{S_{n}}\left(A\left(S_{n}\right)\right)+F_{S_{n}}\left(A\left(S\right)\right)-F_{S}\left(A\left(S\right)\right)\right)\\
	=&\frac{\tau}{n}\left(f^{\prime}\left(A\left(S_{n}\right) ; z_{n}\right)- f^{\prime}\left(A\left(S_{n}\right) ; z_{n}^{\prime}\right)+f^{\prime}\left(A\left(S\right) ; z_{n}^{\prime}\right)-f^{\prime}\left(A\left(S\right) ; z_{n}\right)\right)\\
	&+\frac{(1-\tau)}{n(n-1)}\sum_{i \in[n-1]}\left(\left(g\left(A\left(S_{n}\right) ; z_{i}, z_{n}\right)-g\left(A(S) ; z_{i}, z_{n}\right)\right)+\left(g\left(A\left(S_{n}\right) ; z_{n}, z_{i}\right)-g\left(A(S) ; z_{n}, z_{i}\right)\right)\right. \\
    &+\left.\left(g\left(A(S) ; z_{i}, z_{n}^{\prime}\right)-g\left(A\left(S_{n}\right) ; z_{i}, z_{n}^{\prime}\right)\right)+\left(g\left(A(S) ; z_{n}^{\prime}, z_{i}\right)-g\left(A\left(S_{n}\right) ; z_{n}^{\prime}, z_{i}\right)\right)\right) .
\end{align*}
The proof is completed.\\
\qed
\subsubsection{Proof of Lemma 1}
Let $\mathbf{w}_{S}=A(S)$ and $\mathbf{w}_{S^{\prime}}=A\left(S^{\prime}\right)$. Without loss of generality, we can assume $S^{\prime}=S_n=\left\{z_{1}, \ldots, z_{n-1}, z_{n}^{\prime}\right\}$.
Due to $\mathbf{w}_S$ is the minimizer of $F_{S_{n}}$ and $F_{S}$ is $\sigma$-strongly convex, we know

\begin{equation}\label{strong}
	F_{S}\left(\mathbf{w}_{S^{\prime}}\right)-F_{S}\left(\mathbf{w}_{S}\right) \geq \frac{\sigma}{2}\left\|\mathbf{w}_{S^{\prime}}-\mathbf{w}_{S}\right\|_2^{2}.
\end{equation}
According to
$$
\left|f(\mathbf{w} ; z)-f\left(\mathbf{w}^{\prime} ; z\right)\right| \leq L\left\|\mathbf{w}-\mathbf{w}^{\prime}\right\|_2, \quad \forall z\in \mathcal{Z}, \mathbf{w}, \mathbf{w}^{\prime} \in \mathcal{W},
$$
$$
\left|g(\mathbf{w} ; z, \tilde{z})-g\left(\mathbf{w}^{\prime} ; z, \tilde{z}\right)\right| \leq L\left\|\mathbf{w}-\mathbf{w}^{\prime}\right\|_2, \quad \forall z, \tilde{z} \in \mathcal{Z}, \mathbf{w}, \mathbf{w}^{\prime} \in \mathcal{W},
$$
and Lemma \ref{lem}, we obtain
\begin{align*}
    F_{S}\left(\mathbf{w}_{S^{\prime}}\right)-F_{S}\left(\mathbf{w}_{S}\right) &\leq \frac{2 L\tau\left\|\mathbf{w}_{S^{\prime}}-\mathbf{w}_{S}\right\|_2}{n}+\frac{4(n-1) L(1-\tau)\left\|\mathbf{w}_{S^{\prime}}-\mathbf{w}_{S}\right\|_2}{n(n-1)}\\
    &=\frac{2 L}{n}(2-\tau)\left\|\mathbf{w}_{S^{\prime}}-\mathbf{w}_{S}\right\|_2.
\end{align*}
By combining the above inequality with \eqref{strong}, we have
$$
\left\|\mathbf{w}_{S}-\mathbf{w}_{S^{\prime}}\right\| _2\leq \frac{4 L}{n\sigma}(2-\tau).
$$
With Lemma \ref{lips}, it is easy to get the $\frac{4 L^{2}}{n \sigma}(2-\tau)$-uniform stability of $A$. \\
\qed
\subsubsection{Proof of Lemma 2}
With the reason that $A(S)$ is the minimizer of $F_{S}$, we know that $F_{S}^{\prime}(A(S))=0$. Then, further based on the definition of strong convexity, there holds
\begin{equation}\label{strong convexity}
	R_{S}\left(w^{*}\right)+r\left(w^{*}\right)-R_{S}(A(S))-r(A(S)) \geq \frac{\sigma}{2}\left\|A(S)-w^{*}\right\|_2^{2}.
\end{equation}
From \eqref{pointwise gap}, we know
$$
n\left(R^{point}(A(S))-R^{point}_{S}(A(S))\right) \leq 2 n \gamma+\sum_{i=1} d_{i}.
$$
Similarly, from \citet{lei2020sharper}, we know
$$
    n(n-1)\left(R^{pair}(A(S))-R^{pair}_{S}(A(S))\right) \leq 4 n(n-1) \gamma+\sum_{i, j \in[n]: i \neq j} d_{i, j},
$$
where $\mathbb{E}\left[d_{i, j}\right]=0$. Besides, the proof of Theorem 1 shows that $\mathbb{E}\left[d_{i}\right]=0$. It then follows that
$$
\mathrm{E}[R(A(S))-\operatorname{Rs}(A(S))] \leq 2\gamma(2-\tau) .
$$
We can put the above inequality back into \eqref{strong convexity} to derive

\begin{align*}
	\frac{\sigma}{2} \mathbb{E}\left[\left\|A(S)-\mathbf{w}^{*}\right\|^{2}\right] & \leq \mathbb{E}\left[R_{S}\left(w^{*}\right)+r\left(w^{*}\right)-R_{S}(A(S))-r(A(S))\right] \\
	& \leq \mathbb{E}\left[R_{S}\left(w^{*}\right)+r\left(w^{*}\right)-R(A(S))-r(A(S))\right]+2\gamma(2-\tau) \\
	&=\mathbb{E}\left[R\left(w^{*}\right)+r\left(w^{*}\right)-R(A(S))-r(A(S))\right]+2\gamma(2-\tau) \leq 2\gamma(2-\tau).
\end{align*}

\qed\\
\subsubsection{Proof of Lemma 3}
Because the PPL involves a pointwise part and a pairwise part, some related works of \citet{lei2020sharper} and \citet{pitcan2017note} can be combined here, i.e.,
\begin{align*}
	&\Bigg|\frac{\tau}{n} \sum_{i\in[n]} f\left(\mathbf{w} ; z_{i}\right)+\frac{1-\tau}{n(n-1)} \sum_{i, j \in[n]: i \neq j} g\left(\mathbf{w} ; z_{i}, z_{j}\right)-\tau\mathbb{E}_{z}[f(\mathbf{w} ; z)]-(1-\tau)\mathbb{E}_{z, \bar{z}}[g(\mathbf{w} ; z, \tilde{z})]\Bigg|\\
	\leq&\tau\Bigg|\frac{1}{n} \sum_{i\in[n]} f\left(\mathbf{w} ; z_{i}\right)-\mathbb{E}_{z}[f(\mathbf{w} ; z)]\Bigg|+(1-\tau)\Bigg|\frac{1}{n(n-1)} \sum_{i, j \in[n]: i \neq j} g\left(\mathbf{w} ; z_{i}, z_{j}\right)-\mathbb{E}_{z, \bar{z}}[g(\mathbf{w} ; z, \tilde{z})] \Bigg|\\
	\leq& \frac{2\tau b\mathrm{log}(1/\delta)}{3\lfloor n \rfloor} +\tau\sqrt{\frac{2\theta\mathrm{log}(1/\delta)}{\lfloor n \rfloor}} + \frac{2(1-\tau)b\mathrm{log}(1/\delta)}{3\lfloor n/2 \rfloor} + (1-\tau)\sqrt{\frac{2\theta\mathrm{log}(1/\delta)}{\lfloor n/2 \rfloor}}.
\end{align*}
\\
\qed

\subsubsection{Proof of Theorem 3}
According to Lemma $1$ and Lemma $2$, we deduce $\mathbb{E}_{S}\left[\left\|\mathbf{w}^{*}-A(S)\right\|_2^{2}\right] \leq \frac{16 L^{2}(2-\tau)^{2}}{n \sigma^{2}}$. 
We can further obtain that
\begin{equation}\label{fangsuo}
	\mathbb{E}_{S}\left[\|\mathbf{w}^{*}-A(S)\|_2\right] \leq\left(\mathbb{E}_{S}\left[\left\|\mathbf{w}^{*}-A(S)\right\|_2^{2}\right]\right)^{\frac{1}{2}} \leq \frac{4L(2-\tau)}{\sqrt{n} \sigma}.
\end{equation}
For any $\mathbf{w} \in \mathcal{W}$ and $z, \tilde{z}$, we define
$$
\tilde{f}(\mathbf{w} ; z)=f(\mathbf{w} ; z)-f\left(\mathbf{w}^{*} ; z\right)
$$
and
$$
\tilde{g}(\mathbf{w} ; z, \tilde{z})=g(\mathbf{w} ; z, \tilde{z})-g\left(\mathbf{w}^{*} ; z, \tilde{z}\right). 
$$
With Lemma \ref{lips} and \eqref{fangsuo}, for all $z, \tilde{z} \in \mathcal{Z}$, we can obtain the following inequalities:
\begin{equation*}
	\left|\mathbb{E}_{S}[\tilde{f}(A(S) ; z)]\right|=\left|\mathbb{E}_{S}\left[f(A(S) ; z)-f\left(\mathbf{w}^{*} ; z\right)\right]\right| \leq L \mathbb{E}_{S}\left[\left\|\mathbf{w}^{*}-A(S)\right\|_2\right] \leq \frac{4L^2(2-\tau)}{\sqrt{n} \sigma}
\end{equation*}
and
\begin{equation*}
	\left|\mathbb{E}_{S}[\tilde{g}(A(S) ; z, \bar{z})]\right|=\left|\mathbb{E}_{S}\left[g(A(S) ; z, \tilde{z})-g\left(\mathbf{w}^{*} ; z, \tilde{z}\right)\right]\right| \leq L \mathbb{E}_{S}\left[\left\|\mathbf{w}^{*}-A(S)\right\|_2\right] \leq \frac{4L^2(2-\tau)}{\sqrt{n} \sigma} .
\end{equation*}
We now apply Theorem $1$ with $\gamma=\frac{4 L^{2}}{n \sigma}(2-\tau)$ and $M=\frac{4L^2(2-\tau)}{\sqrt{n} \sigma}$ to show the following inequality  

\begin{align*}
	&\left|\frac{\tau}{n} \sum_{i=1} \tilde{f}\left(A(S) ; z_{i}\right)-\tau\mathbb{E}_{z}[\tilde{f}(A(S) ; z)]+\frac{(1-\tau)}{n(n-1)} \sum_{i \neq j} \tilde{g}\left(A(S) ; z_{i}, z_{j}\right)-(1-\tau)\mathrm{E}_{z, \tilde{z}}[\tilde{g}(A(S) ; z, \tilde{z})]\right|  \\
	\leq& \frac{8 L^{2}}{n \sigma}(2-\tau)^2+e\left(\frac{16 L^{2}}{n \sigma}(2-\tau)(4-3\tau) \sqrt{\log (e / \delta)}+ \frac{96\sqrt{2} L^{2}}{n \sigma}(2-\tau)^2\left\lceil\log _{2}(n)\right\rceil \log (e / \delta)\right)
\end{align*}with probability $1-\delta$.
Therefore,
\begin{align}
	&\left|R_{S}(A(S))-R(A(S))\right| \nonumber\\ \leq&\left|\frac{\tau}{n} \sum_{i=1} f\left(\mathbf{w}^{*} ; z_{i}\right)-\tau\mathbb{E}_{z}\left[f\left(\mathbf{w}^{*} ; z\right)\right]+\frac{(1-\tau)}{n(n-1)} \sum_{i \neq j} g\left(\mathbf{w}^{*} ; z_{i}, z_{j}\right)-(1-\tau)\mathbb{E}_{z, \tilde{z}}\left[g\left(\mathbf{w}^{*} ; z, \tilde{z}\right)\right]\right| \nonumber\\
	&+\frac{8 L^{2}}{n \sigma}(2-\tau)^2+e\left(\frac{16 L^{2}}{n \sigma}(2-\tau)(4-3\tau) \sqrt{\log (e / \delta)}+ \frac{96\sqrt{2} L^{2}}{n \sigma}(2-\tau)^2\left\lceil\log _{2}(n)\right\rceil \log (e / \delta)\right).\label{eq1}
\end{align}
As stated in Lemma $3$, with probability at least $1-\delta$ we have
\begin{align}
	&\Bigg|\frac{\tau}{n} \sum_{i\in[n]} f\left(\mathbf{w}^{*} ; z_{i}\right)+\frac{(1-\tau)}{n(n-1)} \sum_{i, j \in[n]: i \neq j} g\left(\mathbf{w}^{*} ; z_{i}, z_{j}\right)-\tau\mathbb{E}_{z}[f(\mathbf{w}^{*} ; z)]-\mathbb{E}_{z, \tilde{z}}[g(\mathbf{w}^{*} ; z, \tilde{z})]\Bigg| \nonumber\\
	\leq &\frac{2\tau b\log(1/\delta)}{3\lfloor n \rfloor} +\tau\sqrt{\frac{2\theta\log(1/\delta)}{\lfloor n \rfloor}} + \frac{2(1-\tau)b\mathrm{log}(1/\delta)}{3\lfloor n/2 \rfloor} + (1-\tau)\sqrt{\frac{2\theta\mathrm{log}(1/\delta)}{\lfloor n/2 \rfloor}}.\label{eq2}
\end{align}
Combine \eqref{eq1} with \eqref{eq2}, we deduce that with probability $1-\delta$
\begin{align*}
	&\left|R_{S}(A(S))-R(A(S))\right| \\
	\leq&\frac{2\tau b\log(1/\delta)}{3\lfloor n \rfloor} +\tau\sqrt{\frac{2\theta\log(1/\delta)}{\lfloor n \rfloor}} + \frac{2(1-\tau)b\mathrm{log}(1/\delta)}{3\lfloor n/2 \rfloor} + (1-\tau)\sqrt{\frac{2\theta\mathrm{log}(1/\delta)}{\lfloor n/2 \rfloor}} \\
	&+\frac{8 L^{2}}{n \sigma}(2-\tau)^2+e\left(\frac{16 L^{2}}{n \sigma}(2-\tau)(4-3\tau) \sqrt{\log (e / \delta)}+ \frac{96\sqrt{2} L^{2}}{n \sigma}(2-\tau)^2\left\lceil\log _{2}(n)\right\rceil \log (e / \delta)\right).
\end{align*}

The proof is complete.\\
\qed

\subsection{B.4~~~ Proof of Theorem 6}
\subsubsection{Proof of Theorem 4}
Note that $S_{i}$ and $S_{i, j}$ is defined in $(3)$ and $(4)$. 
It is easily known that $\mathbb{E}\left[R^{point}(A(S))\right]=\mathbb{E}\left[R^{point}(A(S_{i}))\right]$ and $\mathbb{E}\left[R^{pair}(A(S))\right]=\mathbb{E}\left[R^{pair}(A(S_{i,j}))\right]$ for all $i,j\in [n]$ with $i\neq j$.
Thus, 
\begin{align*}
	&\mathbb{E}\left[R(A(S))-R_{S}(A(S))\right]\\
	=&\mathbb{E}\left[\tau \left(R^{point}(A(S))-R_{S}^{point}(A(S))\right)+(1-\tau)\left(R^{pair}(A(S))-R_{S}^{pair}(A(S))\right)\right]\\
	=&\frac{\tau}{n} \sum_{i=1}^n \mathbb{E}\left[R^{point}\left(A\left(S_{i}\right)\right)-R^{point}_{S}(A(S))\right]+\frac{(1-\tau)}{n(n-1)} \sum_{i, j \in[n]: i \neq j} \mathbb{E}\left[R^{pair}\left(A\left(S_{i, j}\right)\right)-R^{pair}_{S}(A(S))\right] \\
	=&\frac{\tau}{n} \sum_{i=1}^n \mathbb{E}\left[f\left(A\left(S_{i}\right) ; z_{i}\right)-f\left(A(S) ; z_{i}\right)\right]+\frac{(1-\tau)}{n(n-1)} \sum_{i, j \in[n]: i \neq j} \mathbb{E}\left[g\left(A\left(S_{i, j}\right) ; z_{i}, z_{j}\right)-g\left(A(S) ; z_{i}, z_{j}\right)\right] \leq \gamma.
\end{align*}
The proof is complete.\\
\qed
\subsubsection{Proof of Lemma 4}
According to the Lemma $2.1$ of \citet{srebro2010smoothness}, we know that the self-bounding property for non-negative and smooth functions, i.e.,
$$
\left\|f^{\prime}(\mathbf{w})\right\|_2^{2} \leq 2 \beta f(\mathbf{w}), \quad \forall \mathbf{w} \in \mathcal{W},
$$
$$
\left\|g^{\prime}(\mathbf{w})\right\|_2^{2} \leq 2 \beta g(\mathbf{w}), \quad \forall \mathbf{w} \in \mathcal{W},
$$
and the following inequalities are also satisfied
$$
f(\mathbf{w}) \leq f\left(\mathbf{w}^{\prime}\right)+\left\langle f^{\prime}\left(\mathbf{w}^{\prime}\right), \mathbf{w}-\mathbf{w}^{\prime}\right\rangle+\frac{\beta\left\|\mathbf{w}-\mathbf{w}^{\prime}\right\|_2^{2}}{2}, \quad \forall \mathbf{w}, \mathbf{w}^{\prime} \in \mathcal{W},
$$
$$
g(\mathbf{w}) \leq g\left(\mathbf{w}^{\prime}\right)+\left\langle g^{\prime}\left(\mathbf{w}^{\prime}\right), \mathbf{w}-\mathbf{w}^{\prime}\right\rangle+\frac{\beta\left\|\mathbf{w}-\mathbf{w}^{\prime}\right\|_2^{2}}{2}, \quad \forall \mathbf{w}, \mathbf{w}^{\prime} \in \mathcal{W}.
$$
Based on the above conditions and the Cauchy-Schwartz inequality, we deduce that
\begin{align*}
& f\left(A\left(S_{i}\right) ; z_{i}\right)-f\left(A(S) ; z_{i} \right) \\
\leq& \left\langle f^{\prime}\left(A(S) ; z_{i}\right), A\left(S_{i}\right)-A(S)\right\rangle+\frac{\beta}{2}\left\|A\left(S_{i}\right)-A(S)\right\|_2^{2} \\
\leq& \left\|f^{\prime}\left(A(S) ; z_{i}\right)\right\|_2\left\|A\left(S_{i}\right)-A(S)\right\|_2+\frac{\beta}{2}\left\|A\left(S_{i}\right)-A(S)\right\|_2^{2} \\
\leq & \frac{\left\|f^{\prime}\left(A(S) ; z_{i}\right)\right\|_2^{2}}{2 \epsilon}+\frac{\epsilon+\beta}{2}\left\|A\left(S_{i}\right)-A(S)\right\|_2^{2} \leq \frac{\beta f\left(A(S) ; z_{i}\right)}{\epsilon}+\frac{\epsilon+\beta}{2}\left\|A\left(S_{i}\right)-A(S)\right\|_2^{2}.
\end{align*}
Then, we apply the above inequality to Theorem 5 to obtain
\begin{align}
&\mathbb{E}\left[R^{point}(A(S))-R_{S}^{point}(A(S))\right] \nonumber\\
\leq &\frac{\beta}{\epsilon n} \sum_{i=1}^n \mathbb{E}\left[f\left(A(S) ; z_{i}\right)\right]+\frac{\epsilon+\beta}{2 n} \sum_{i=1}^n \mathbb{E}\left[\left\|A\left(S_{i}\right)-A(S)\right\|_2^{2}\right] \nonumber\\
=&\frac{\beta \mathbb{E}\left[R_{S}^{point}(A(S))\right]}{\epsilon}+\frac{\epsilon+\beta}{2 n} \sum_{i=1}^n \mathbb{E}\left[\left\|A\left(S_{i}\right)-A(S)\right\|_2^{2}\right] .\label{point}
\end{align}
From the Lemma D.1 of \citet{lei2020sharper}, we know
\begin{equation}\label{pair}
   \mathbb{E}\left[R^{pair}(A(S))-R_{S}^{pair}(A(S))\right] \leq \frac{\beta \mathbb{E}\left[R_{S}^{pair}(A(S))\right]}{\epsilon}+\frac{2(\epsilon+\beta)}{n} \sum_{i=1}^n \mathbb{E}\left[\left\|A\left(S_{i}\right)-A(S)\right\|_2^{2}\right].
\end{equation}
Combining \eqref{point} and \eqref{pair}, we have
\begin{align*}
    &\mathbb{E}\left[R(A(S))-R_{S}(A(S))\right]\\
    \leq& \frac{\beta\tau \mathbb{E}\left[R_{S}^{point}(A(S))\right]}{\epsilon}+\frac{\beta(1-\tau) \mathbb{E}\left[R_{S}^{pair}(A(S))\right]}{\epsilon}+\frac{(\epsilon+\beta)}{n}(2-\frac{3}{2}\tau) \sum_{i=1}^n \mathbb{E}\left[\left\|A\left(S_{i}\right)-A(S)\right\|_2^{2}\right].
\end{align*}
The proof is completed.\\
\qed
\subsubsection{Proof of Theorem 6}
Due to Lemma \ref{lem} and the $\beta$-smoothness of $f$ and $g$, we deduce that, for any $k\in [n]$,
\begin{align*}
    F_{S}\left(A\left(S_{k}\right)\right)-F_{S}(A(S)) \leq&\frac{\tau}{n}\left(\langle f^{\prime}\left(A(S) ; z_{k}\right)-f^{\prime}\left(A\left(S_{k}\right) ; z_{k}^{\prime}\right) , A\left(S_{k}\right)-A(S)\rangle+ \beta\left\|A\left(S_{k}\right)-A(S)\right\|_2^{2}\right)\\
    &+\frac{(1-\tau)}{n(n-1)}\sum_{i \in[n]: i \neq k}\left(\left\langle g^{\prime}\left(A(S) ; z_{i}, z_{k}\right)+g^{\prime}\left(A(S) ; z_{k}, z_{i}\right)-g^{\prime}\left(A\left(S_{k}\right) ; z_{i}, z_{k}^{\prime}\right)\right.\right.\\
    & \left. -g^{\prime}\left(A\left(S_{k}\right) ; z_{k}^{\prime}, z_{i}\right) , A\left(S_{k}\right)-A(S)\right\rangle+\frac{4 \beta\left\|A\left(S_{k}\right)-A(S)\right\|_2^{2}}{2}) .
\end{align*}
Using the Cauchy-Schwartz inequality, we obtain
\begin{align*}
    F_{S}\left(A\left(S_{k}\right)\right)-F_{S}(A(S)) \leq& \frac{\tau}{n}\left(\left(\left\|f^{\prime}\left(A(S) ; z_{k}\right)\right\|_2+\left\|f^{\prime}\left(A\left(S_{k}\right) ; z_{k}^{\prime}\right)\right\|_2\right)\left\|A\left(S_{k}\right)-A(S)\right\|_2+ \beta\left\|A\left(S_{k}\right)-A(S)\right\|_2^{2}\right)\\
    &+\frac{(1-\tau)}{n(n-1)}(\sum_{i \in[n]: i \neq k}\left(\left\|g^{\prime}\left(A(S) ; z_{i}, z_{k}\right)\right\|_2+\left\|g^{\prime}\left(A(S) ; z_{k}, z_{i}\right)\right\|_2+\left\|g^{\prime}\left(A\left(S_{k}\right) ; z_{i}, z_{k}^{\prime}\right)\right\|_2\right.\\
    &+\left.\left\|g^{\prime}\left(A\left(S_{k}\right) ; z_{k}^{\prime}, z_{i}\right)\right\|_2\right)\left\|A\left(S_{k}\right)-A(S)\right\|_2+2 \beta(n-1)\left\|A\left(S_{k}\right)-A(S)\right\|_2^{2}) .
\end{align*}
Based on the self-bounding property and strong-convexity property, we verify that 

\begin{align*}
    \frac{\sigma \left\|A\left(S_{k}\right)-A(S)\right\|_2^{2}}{2} \leq& \frac{\tau}{n}\left(\sqrt{2 \beta} \left(\sqrt{f\left(A(S) ;z_{k}\right)}+\sqrt{f\left(A\left(S_{k}\right) ; z_{k}^{\prime}\right)}\right)\left\|A\left(S_{k}\right)-A(S)\right\|_2+ \beta\left\|A\left(S_{k}\right)-A(S)\right\|_2^{2}\right)\\
    &+\frac{(1-\tau)}{n(n-1)}(\sqrt{2 \beta} \sum_{i \in[n]: i \neq k}\left(\sqrt{\ell\left(A(S) ; z_{i}, z_{k}\right)}+\sqrt{\ell\left(A(S) ; z_{k}, z_{i}\right)}+\sqrt{\ell\left(A\left(S_{k}\right) ; z_{i}, z_{k}^{\prime}\right)}\right. \\
    &\left.+\sqrt{\ell\left(A\left(S_{k}\right) ; z_{k}^{\prime}, z_{i}\right)}\right)\left\|A\left(S_{k}\right)-A(S)\right\|_2+2 \beta(n-1)\left\|A\left(S_{k}\right)-A(S)\right\|_2^{2})
\end{align*}
and
\begin{align*}
    \frac{\sigma \left\|A\left(S_{k}\right)-A(S)\right\|_2}{2} \leq& \frac{\tau}{n}\left(\sqrt{2 \beta} \left(\sqrt{f\left(A(S) ;z_{k}\right)}+\sqrt{f\left(A\left(S_{k}\right) ; z_{k}^{\prime}\right)}\right)+ \beta\left\|A\left(S_{k}\right)-A(S)\right\|_2\right)\\
    &+\frac{(1-\tau)}{n(n-1)}(\sqrt{2 \beta} \sum_{i \in[n]: i \neq k}\left(\sqrt{g\left(A(S) ; z_{i}, z_{k}\right)}+\sqrt{g\left(A(S) ; z_{k}, z_{i}\right)}\right. \\
    &\left.+\sqrt{g\left(A\left(S_{k}\right) ; z_{i}, z_{k}^{\prime}\right)}+\sqrt{g\left(A\left(S_{k}\right) ; z_{k}^{\prime}, z_{i}\right)}\right)+2 \beta(n-1)\left\|A\left(S_{k}\right)-A(S)\right\|_2).
\end{align*}
When $ \beta \leq \sigma n / 4(2-\tau)$,
\begin{align*}
    &\frac{\sigma \left\|A\left(S_{k}\right)-A(S)\right\|_2}{4} \\
    \leq &\frac{\tau}{n}\left(\sqrt{2 \beta} \left(\sqrt{f\left(A(S) ;z_{k}\right)}+\sqrt{f\left(A\left(S_{k}\right) ; z_{k}^{\prime}\right)}\right)\right)\\
    &+\frac{(1-\tau)}{n(n-1)}\sqrt{2 \beta} \sum_{i \in[n]: i \neq k}\left(\sqrt{g\left(A(S) ; z_{i}, z_{k}\right)}+\sqrt{g\left(A(S) ; z_{k}, z_{i}\right)}+\sqrt{g\left(A\left(S_{k}\right) ; z_{i}, z_{k}^{\prime}\right)}+\sqrt{g\left(A\left(S_{k}\right) ; z_{k}^{\prime}, z_{i}\right)}\right).
\end{align*}
We multiply both sides of the above inequality by $n(n-1)$ to derive
\begin{align*}
    &\frac{\sigma n(n-1)\left\|A\left(S_{k}\right)-A(S)\right\|_2}{4} \\
    \leq &\tau (n-1)\left(\sqrt{2 \beta} \left(\sqrt{f\left(A(S) ;z_{k}\right)}+\sqrt{f\left(A\left(S_{k}\right) ; z_{k}^{\prime}\right)}\right)\right)\\
    &+(1-\tau)\sqrt{2 \beta} \sum_{i \in[n]: i \neq k}\left(\sqrt{g\left(A(S) ; z_{i}, z_{k}\right)}+\sqrt{g\left(A(S) ; z_{k}, z_{i}\right)}+\sqrt{g\left(A\left(S_{k}\right) ; z_{i}, z_{k}^{\prime}\right)}+\sqrt{g\left(A\left(S_{k}\right) ; z_{k}^{\prime}, z_{i}\right)}\right),
\end{align*}
and further square, that is,
\begin{align*}
    &\frac{\sigma^2 n^2(n-1)^2\left\|A\left(S_{k}\right)-A(S)\right\|_2^2}{16}
    \\
    \leq& 2\beta(n-1)\sum_{i \in[n]: i \neq k}\left((1-\tau)\left(\sqrt{g\left(A(S) ; z_{i}, z_{k}\right)}+\sqrt{g\left(A(S) ; z_{k}, z_{i}\right)}\right.\right.\\
    &\left.\left.+\sqrt{g\left(A\left(S_{k}\right) ; z_{i}, z_{k}^{\prime}\right)}+\sqrt{g\left(A\left(S_{k}\right) ; z_{k}^{\prime}, z_{i}\right)}\right)+\tau\left(\sqrt{f\left(A(S) ;z_{k}\right)}+\sqrt{f\left(A\left(S_{k}\right) ; z_{k}^{\prime}\right)}\right)\right)^2.
\end{align*}
Then we use Cauchy-Schwartz inequality to derive
\begin{align*}
    &\sigma^2 n^2(n-1)\left\|A\left(S_{k}\right)-A(S)\right\|_2^2\\
    \leq  &192\beta 
    \sum_{i \in[n]: i \neq k}\left((1-\tau)^2\left(g\left(A(S) ; z_{i}, z_{k}\right)+g\left(A(S) ; z_{k}, z_{i}\right)
    +g\left(A\left(S_{k}\right) ; z_{i}, z_{k}^{\prime}\right)+g\left(A\left(S_{k}\right) ; z_{k}^{\prime}, z_{i}\right)\right)\right.\\
    &+\left.\tau^2\left(f\left(A(S) ;z_{k}\right)+f\left(A\left(S_{k}\right) ; z_{k}^{\prime}\right)\right)\right).
\end{align*}
We take a summation from $k=1$ to $n$ to obtain
\begin{align*}
    &\sigma^2 n^2(n-1)\sum_{k=1}^{n}\left\|A\left(S_{k}\right)-A(S)\right\|_2^2\\ 
    \leq  &192\beta 
    \sum_{i, k \in[n]: i \neq k}[(1-\tau)^2\left(g\left(A(S) ; z_{i}, z_{k}\right)+g\left(A(S) ; z_{k}, z_{i}\right)
    +g\left(A\left(S_{k}\right) ; z_{i}, z_{k}^{\prime}\right)+g\left(A\left(S_{k}\right) ; z_{k}^{\prime}, z_{i}\right)\right)\\
    &+\tau^2\left(f\left(A(S) ;z_{k}\right)+f\left(A\left(S_{k}\right) ; z_{k}^{\prime}\right)\right)].
\end{align*}
Due to the symmetry, it is easily known that
\begin{equation*}
\sigma^2 n\sum_{k=1}^{n}\left\|A\left(S_{k}\right)-A(S)\right\|_2^2 \leq 384\tau^2\beta\mathbb{E}\left[R_{S}^{point}(A(S))\right]+768 (1-\tau)^2\beta \mathbb{E}\left[R_{S}^{pair}(A(S))\right].
\end{equation*}
With Lemma $4$, we derive the following inequality for all $\epsilon>0$
\begin{align*}
    &\mathbb{E}\left[R(A(S))-R_{S}(A(S))\right]\\
    \leq &\frac{\beta\tau \mathbb{E}\left[R_{S}^{point}(A(S))\right]}{\epsilon}+\frac{\beta(1-\tau) \mathbb{E}\left[R_{S}^{pair}(A(S))\right]}{\epsilon}+\frac{(\epsilon+\beta)}{n}\left(2-\frac{3}{2}\tau\right) \sum_{i=1}^{n} \mathbb{E}\left[\left\|A\left(S_{i}\right)-A(S)\right\|_2^{2}\right]\\
    \leq&\frac{\beta\tau \mathbb{E}\left[R_{S}^{point}(A(S))\right]}{\epsilon}+\frac{384\tau^2(\epsilon+\beta)\beta}{\sigma^2n^2}\left(2-\frac{3}{2}\tau\right)\mathbb{E}\left[R_{S}^{point}(A(S))\right]\\
    &+\frac{\beta(1-\tau) \mathbb{E}\left[R_{S}^{pair}(A(S))\right]}{\epsilon}+\frac{768(1-\tau)^2(\epsilon+\beta)\beta}{\sigma^2n^2}\left(2-\frac{3}{2}\tau\right)\mathbb{E}\left[R_{S}^{pair}(A(S))\right].
\end{align*}
In addition, from the definition of $A(S)$, we get 
\begin{align*}
\mathbb{E}[F(A(S))]-F\left(\mathbf{w}^{*}\right) &=\mathbb{E}\left[F(A(S))-F_{S}(A(S))\right]+\mathbb{E}\left[F_{S}(A(S))-F_{S}\left(\mathbf{w}^{*}\right)\right] \\
& \leq \mathbb{E}\left[F(A(S))-F_{S}(A(S))\right]=\mathbb{E}\left[R(A(S))-R_{S}(A(S))\right].
\end{align*}
Thus,
\begin{align*}
    &\mathbb{E}[F(A(S))]-F\left(\mathbf{w}^{*}\right)\\
    \leq&\frac{\beta\tau \mathbb{E}\left[R_{S}^{point}(A(S))\right]}{\epsilon}+\frac{384\tau^2(\epsilon+\beta)\beta}{\sigma^2n^2}\left(2-\frac{3}{2}\tau\right)\mathbb{E}\left[R_{S}^{point}(A(S))\right]\\
    &+\frac{\beta(1-\tau) \mathbb{E}\left[R_{S}^{pair}(A(S))\right]}{\epsilon}+\frac{768(1-\tau)^2(\epsilon+\beta)\beta}{\sigma^2n^2}\left(2-\frac{3}{2}\tau\right)\mathbb{E}\left[R_{S}^{pair}(A(S))\right].
\end{align*}
The desired result is proved. \qed

\end{document}